\documentclass[10pt,twocolumn,letterpaper]{article}

\usepackage{amsfonts,amsmath,amssymb}
\usepackage{array}
\usepackage{bbm}
\usepackage{blkarray}
\usepackage{booktabs}
\usepackage{color}
\usepackage{comment}
\usepackage{dsfont}
\usepackage{enumitem}
\usepackage{epsfig}
\usepackage[T1]{fontenc}
\usepackage{graphicx}
\usepackage{makecell}
\usepackage{multirow}
\usepackage{siunitx}
\usepackage{tabularx}
\usepackage{times}
\usepackage{xspace}
\usepackage{wrapfig}
\usepackage{ICCV/iccv}

\let\originalparagraph\paragraph
\renewcommand{\paragraph}[2][.]{\originalparagraph{#2#1}}

\newcolumntype{L}{>{\raggedright\arraybackslash}X}
\newcolumntype{C}{>{\centering\arraybackslash}X}

\DeclareMathOperator*{\argmax}{argmax}

\makeatletter
\DeclareRobustCommand\onedot{\futurelet\@let@token\@onedot}
\def\@onedot{\ifx\@let@token.\else.\null\fi\xspace}

\def\etal{\emph{et al}\onedot}
\makeatother

\graphicspath{{figures/}}

\usepackage[pagebackref=true,breaklinks=true,letterpaper=true,colorlinks,bookmarks=false]{hyperref}

\iccvfinalcopy
\def\iccvPaperID{6977}

\ificcvfinal\fi

\begin{document}

\title{CPFN: Cascaded Primitive Fitting Networks for High-Resolution Point Clouds}

\author{Eric-Tuan Lê\textsuperscript{1}\thanks{This work was partly done when E.~Lê interned and M.~Sung worked at Adobe Research.} $\;$
Minhyuk Sung\textsuperscript{2}\footnotemark[1] $\;$
Duygu Ceylan\textsuperscript{3} $\;$
Radomir Mech\textsuperscript{3} $\;$
Tamy Boubekeur\textsuperscript{3} $\;$
Niloy J. Mitra\textsuperscript{1,3} \\
\textsuperscript{1}University College London $\quad$
\textsuperscript{2}KAIST $\quad$
\textsuperscript{3}Adobe Research\\
}

\maketitle
\ificcvfinal\thispagestyle{empty}\fi

\begin{abstract}
\vspace{-5pt}

Representing human-made objects as a collection of base primitives has a long history in computer vision and reverse engineering. In the case of high-resolution point cloud scans, the challenge is to be able to detect both large primitives as well as those explaining the detailed parts. 
While the classical RANSAC approach requires case-specific parameter tuning, state-of-the-art networks are limited by memory consumption of their backbone modules such as PointNet++~\cite{PointNet++}, and hence fail to detect the fine-scale primitives.
We present Cascaded Primitive Fitting Networks~(CPFN) that relies on an adaptive patch sampling network to assemble detection results of global and local primitive detection networks. As a key enabler, we present a merging formulation that dynamically aggregates the primitives across global and local scales. 
Our evaluation demonstrates that CPFN improves the state-of-the-art SPFN performance by $13-14\%$ on high-resolution point cloud datasets and specifically improves the detection of fine-scale primitives by $20-22\%$.
Our code is available at:
\href{https://github.com/erictuanle/CPFN}{https://github.com/erictuanle/CPFN}

\vspace{-5pt}
\end{abstract}

\section{Introduction}
\vspace{-3pt}

Representing 3D shapes with a compact set of atomic primitives is a well-established idea that has been evolved over the decades~\cite{Binford:1971,Marr:1978}. While the idea has been mainly exploited for machine perception in a way to parse objects, most human-made objects are indeed modeled as a composition of geometric primitives. In CAD, modeling techniques such as Constructive Solid Geometry (CSG)~\cite{CSG} or building a binary tree of simple primitives, have been conventional practices. Hence, for scanned data of human-made objects, converting them into a form to reflect how they were modeled is important not only for the perception but also for enabling editing capabilities in downstream applications. The problem of precisely fitting primitives to the input scan is, however, more challenging than coarsely parsing and abstracting the shape.

For such a model fitting problem, RANSAC~\cite{Fischler:1981} is the de facto standard technique in computer vision. The algorithm of Schnabel~\etal~\cite{Schnabel:2007} or Li~\etal\cite{li_globFit_sigg11} which iteratively runs RANSAC to find fitting primitives has been implemented in popular geometry processing libraries such as CGAL~\cite{CgalRANSAC} and applied to solve the primitive fitting problem with many real scan data. However, such an unsupervised approach often suffers from the combinatorial complexity nature of the problem. From an optimization perspective, different primitive configurations can potentially result in similarly small fitting errors, although the iterative heuristic algorithm cannot take into account all the possible configurations. Furthermore, an undesired set of primitives can even result in a smaller fitting error due to noise in the input. While the RANSAC-based approach deals with the noise to some extent with some threshold parameters, the input-specific parameter tuning requires substantial manual effort.

%
\begin{figure}[t]
    \centering
    \includegraphics[width=\linewidth]{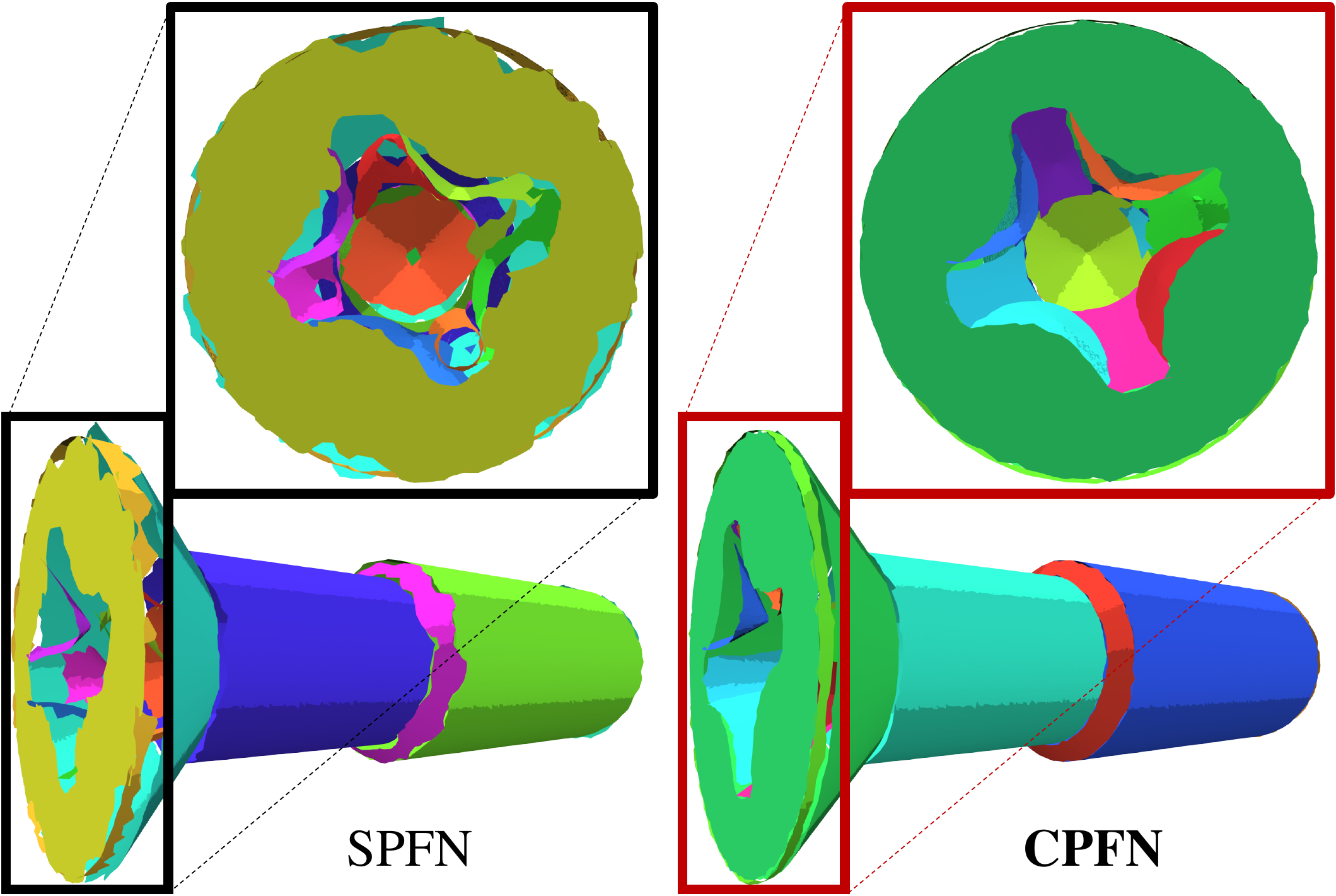}
    \caption{A side-by-side comparison between SPFN~\cite{SPFN} and our CPFN. Our cascaded networks are designed to accurately detect and fit small primitives in a high-resolution point cloud.}
    \label{fig:teaser}
    \vspace{-\baselineskip}
\end{figure}
%

To tackle the challenge, Li~\etal~\cite{SPFN} recently proposed a \emph{supervised} framework called SPFN that learns the best configuration of primitives for each 3D scan from a large collection of CAD data. Instead of directly regressing the primitive parameters, their network employs PointNet++~\cite{PointNet++} as an encoder of the input point cloud and predicts per-point information, including association from a point to a primitive, primitive type, and surface normal. A subsequent differentiable module computes the best primitive parameters minimizing the fitting error through an analytic formulation.

While SPFN~\cite{SPFN} demonstrated successful results, the challenge remains in handling \emph{high-resolution} data. Even affordable 3D scanners are now capable of capturing local geometric details with high-resolution (e.g., point sets with $100$k+ points). However, efficiently processing the high-resolution 3D data in neural networks raises a memory limit issue with consumer GPUs. Even with a simple point cloud processing architecture such as PointNet~\cite{PointNet}, the order of $10$k points is the limit in training, whereas scans may include points in the order of $100$k to $1$M. Downsampling the input point cloud results in information loss for fine-scale details and thus fails to fit small primitives (see Figure~\ref{fig:teaser} for example results with missed features by SPFN on typical high-resolution scans). 

In this work, we propose a novel framework named Cascaded Primitive Fitting Networks (CPFN), which is particularly developed to capture local details in scans and fit small primitives. Our framework cascades two fitting networks: one for processing the \emph{entire} input point cloud, and the other for processing \emph{local} patches of the input. Both of them are SPFNs~\cite{SPFN} but trained separately with global/local input data. 
Our design breaks the problem into three steps: first, adaptively sampling patches in regions of small details; suitably regressing primitives in (detected) regions of fine details; and merging the global and local primitives to get a multi-scale output.

The framework includes a \textit{patch selection network} trained to detect regions with small primitives so that local patches fed to the \textit{local fitting network} can be sampled in those regions at test time. Our key idea is in the merging algorithm that aggregates the per-point outputs of both networks and produces the final fitted primitives. The merging process is formulated as a binary program, although we empirically found that a Hungarian algorithm~\cite{Kuhn:1955} can obtain a near-optimum solution in most cases. In experiments, we demonstrate that our cascaded networks outperform a single SPFN trained with downsampled point clouds in fitting primitives in all scales with a performance boost of $13-14\%$. The improvement reaches $20-22\%$ for smaller primitives. We also show that the fitting performance of the local fitting network can be improved when it takes global contextual information of the entire input point cloud from the global fitting network.

%
In summary, our key contributions are as follows.
\vspace{-3pt}
\begin{itemize}
    \setlength\itemsep{-0.2em}
    \item We propose CPFN, a primitive fitting framework leveraging two cascaded networks to adaptively detect both small and large primitives.
    \item Our merging algorithm ensembles the per-point information predicted by the two networks efficiently and produces the final fitted primitives.
    \item Our experiments demonstrate that the performance of the local fitting network benefits from feeding contextual information learned by the global fitting network.
\end{itemize}
%
\section{Related Work}
\vspace{-3pt}

We review previous work leveraging neural networks in primitive fitting as well as recent work on processing high-resolution point clouds using neural networks. We refer the reader to the recent survey~\cite{Kaiser:2018} for a discussion of classical approaches, particularly RANSAC-based methods.

\vspace{-3pt}
\paragraph{Neural Geometric Primitive Fitting}
Neural-network-based approaches have been widely investigated in decomposing 3D shapes into various types of primitives. To our knowledge, Tulsiani~\etal~\cite{Tulsiani:2017}  and Zou~\etal~\cite{Zou:2017} were the first proposing neural decomposition. They presented networks fitting cuboids to the input 3D shape represented as either voxels or a depth map. Subsequent works have extended this idea. For instance, Sun~\etal~\cite{Sun:2019} created an architecture predicting a hierarchical structure of cuboids; Smirnov~\etal~\cite{Smirnov:2020} suggested a new loss function based on distance fields and fitted rounded cuboids; while Lin~\etal~\cite{Lin:2020} introduced a reinforcement-learning-based approach fitting connected trapezoid boxes sequentially so that the final output becomes a scaffold mesh. Follow up work has also focused on fitting different types of primitives. As a generalization of cuboids, Paschalidou~\cite{Paschalidou:2019} used superquadrics as geometric primitives;  Gadelha~\etal~\cite{shapehandles} explored the use of sphere meshes in the context of learning a generative model of man-made shapes; while Chen~\etal~\cite{BSP-Net} and Deng~\etal~\cite{Deng:2020} concurrently proposed to represent input shapes as a convex set of planes that recursively partition the space. Genova~\etal~\cite{Genova:2019,Genova:2020} used implicit representations, namely Gaussians, to explore locality.
The primitive types used in these works, however, have limited expressibility, and hence these works mostly focused on abstracting the input shapes with a coarse fitting.

Some notable efforts considered multiple primitive types in the fitting. Sharma~\etal~\cite{Sharma:2018} and Kania~\etal~\cite{UCSG-Net} introduced networks predicting a CSG structure from a raw geometry with various types of primitives. However, the precision of fitting was limited since the loss was defined with occupancies in a low-resolution voxel grid. Li~\etal~\cite{SPFN} proposed SPFN, a framework for more precisely fitting multiple primitives including plane, sphere, cylinder, and cone. It was further extended by Sharma~\etal~\cite{Sharma:2020} to fit B-spline patches as well. While showing impressive results, both approaches employ PointNet++~\cite{PointNet++}, an off-the-shelf architecture, for encoding the input point cloud and thus are limited by the point cloud size (e.g., order of 8k points). Our method extends SPFN in a novel cascaded framework that can fit primitives at various scales to a high-resolution point cloud (e.g., order of 128k points).

%
\begin{figure*}[t!]
    \centering
    \includegraphics[width=\textwidth]{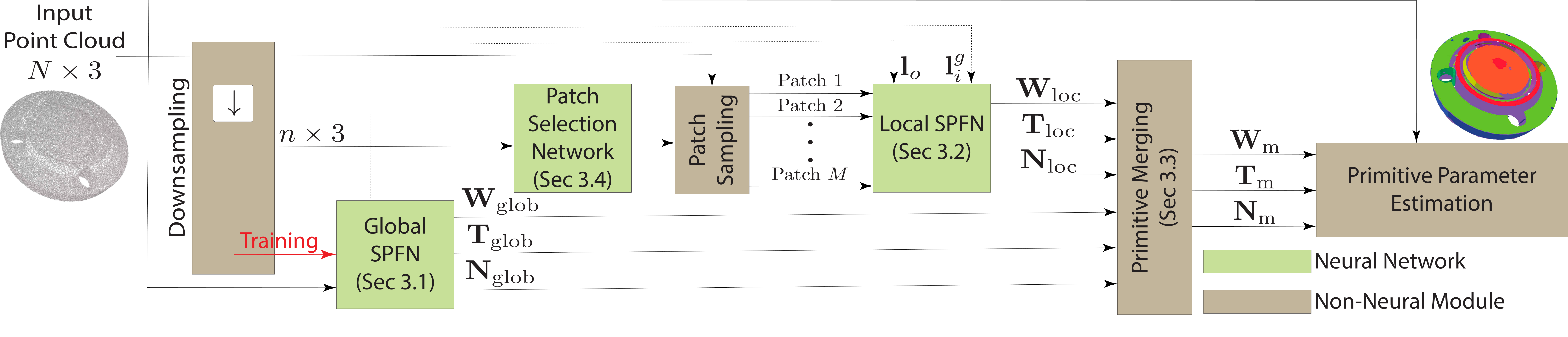}
    \vspace{-1\baselineskip}
    \caption{A diagram of CPFN. CPFN includes two SPFNs~\cite{SPFN}: one for the entire object and the other for local patches. Contextual information is fed from the global SPFN~(Section~\ref{sec:globalSPFN}) to the local SPFN~(Section~\ref{sec:patchPrediction}). Patch Selection Network~(Section~\ref{sec:patchSelection}) takes a downsampled point cloud as input and determines where the local patches should be sampled at test time. The per-point predictions from both SPFNs are integrated in the merging step (Section~\ref{sec:patchMerging}).}
    \label{fig:cpfn_pipeline}
    \vspace{-1\baselineskip}
\end{figure*}
%

\vspace{-3pt}
\paragraph{Neural Networks for High-Resolution Point Clouds}
Recent work has focused on processing high-resolution 3D data as input in neural networks, particularly to handle large-scale indoor/outdoor scans and detect and/or segment objects. In terms of voxel representations, OctNet~\cite{OctNet} and SparseConvNet~\cite{Graham:2018} are examples introducing efficient architectures to avoid computation in empty space. Particularly, SparseConvNet has shown the best performance in 3D indoor scene segmentation, as demonstrated by Han~\etal~\cite{Han:2020}.
The others introduced networks that internally voxelize input point clouds to utilize 3D convolutions~\cite{Rethage:2018,Chen:2019}. Although these architectures perform well in scene segmentation, they are not suitable for fitting problems since the voxelization causes significant discretization errors. As an alternative to voxelization, Tatarchenko~\etal~\cite{Tatarchenko:2018} proposed to exploit 2D convolutions by projecting points in a local region to a tangent plane. However, this architecture is sensitive to errors in surface normal estimation. Other previous work on neural point cloud processing proposed to either cluster points and generate object candidates for instance segmentation~\cite{Landrieu:2018,Chen:2020,Jiang:2020} or concatenate hierarchical downsampling and upsampling modules as an encoder-decoder architecture for semantic segmentation~\cite{Hu:2020,Xu:2020}. Neither of these approaches are directly applicable to our problem since we need to jointly solve for both the semantic and instance segmentations. Our method operates directly at the point level and ensembles fitting results obtained from both the coarse global point cloud as well as high-resolution local patches in a novel merging step to solve the joint segmentation problem.
\section{Method}
\vspace{-3pt}

In this section we describe our cascaded primitive fitting networks. Given an input point cloud with $N$ points (where $N$ is $128$k in our experiments), our networks operate at two levels, namely global and local. We utilize a global primitive fitting network, SPFN~\cite{SPFN}, trained on downsampled versions of the input point clouds (i.e., trained on point clouds of size $n$ where $n \ll N$) due to high memory footprints of point cloud processing backbone modules such as PointNet++. While the trained network can be tested on the original point cloud at inference time, it is likely to miss the small primitives representing fine scale details since they are lost in the downsampling process during training. Hence, we train an additional version of SPFN that operates on local patches of the high resolution point clouds (Section~\ref{sec:patchPrediction}). Given the local predictions for patches and the global predictions for the rest of the point cloud, the core of our method is a novel merging step (Section~\ref{sec:patchMerging}) that consolidates all the predictions. In order to ensure that the capacity of the local network is utilized to learn the prediction of small primitives, at training stage we utilize a smart strategy to select the training patches from regions of the point cloud that contain such primitives. At inference time, we utilize a patch selection network (Section~\ref{sec:patchSelection}) that predicts the regions that are likely to contain small primitives and thus should be processed with the local network. In the following, we first provide a short summary of the SPFN architecture and discuss the different stages of our method in detail as presented in Figure~\ref{fig:cpfn_pipeline}.
 
\subsection{Supervised Primitive Fitting Network~\cite{SPFN}}
\label{sec:globalSPFN}
\vspace{-3pt}

SPFN~\cite{SPFN}, the Supervised Primitive Fitting Network, is an end-to-end network trained to detect the set of primitives (specifically, planes, spheres, cylinders, and cones) of an input 3D point cloud. SPFN first predicts three per-point properties; a segment label $\mathbf{W}$, normal $\mathbf{N}$, and primitive type $\mathbf{T}$. Given such predictions, the actual primitive parameters are estimated in a differentiable manner. Ground truth per-point primitive associations and primitive fitting errors are used as strong supervision. Specifically, the loss is composed of multiple terms: (i)~segmentation loss $\mathcal{L}_{\text{seg}}$, (ii)~normal loss $\mathcal{L}_{\text{norm}}$, (iii)~primitive type loss $\mathcal{L}_{\text{type}}$, (iv)~residual loss $\mathcal{L}_{\text{res}}$, i.e., fitting loss, and (v)~axis loss $\mathcal{L}_{\text{axis}}$ that considers the normal of a plane or the axis of a cylinder or a cone:
%
\begin{align}
\vspace{-3pt}
  \mathcal{L} = \mathcal{L}_{\text{seg}} + \mathcal{L}_{\text{norm}} + \mathcal{L}_{\text{type}} + \mathcal{L}_{\text{res}} + \mathcal{L}_{\text{axis}}.
\vspace{-3pt}
\end{align}
%

When computing the loss, the predicted primitives are first mapped to the ground truth primitives using the Hungarian matching algorithm to find the pairs of primitives that maximize the intersection over union across the matched primitives. Per-point and per-primitive losses are computed based on this correspondence. We refer the reader to the original paper~\cite{SPFN} for details.

Our pipeline leverages the SPFN framework as it is in the global branch. In addition, we train a local version of SPFN that operates on local patches sampled from the high resolution point cloud. We provide additional contextual information to the local SPFN to boost the performance of the local predictions as we will discuss in Section~\ref{sec:patchPrediction}.

\subsection{Local SPFN}
\label{sec:patchPrediction}
\vspace{-3pt}

One  of  the  key  components  of  our  pipeline  is  the  local SPFN module, which aims to predict small primitives in fine-scale regions of the input point clouds. Given a patch sampled on the input point cloud, local SPFN predicts the same per-point features, i.e. the segment label $\mathbf{W}_p$, normal $\mathbf{N}_p$, and primitive type $\mathbf{T}_p$ as the global SPFN. While we keep the architecture of the original SPFN fixed, we provide additional global contextual information as input.
Specifically, given the point cloud representing an object $o$, we first extract a latent vector, $\mathbf{l}_o$, for the entire point cloud using the global SPFN.
Similarly, for each patch $i$, we extract patch features, $\mathbf{l}_i^g$, that we obtain for the seed point of the patch using the global SPFN. We concatenate both the object and patch features obtained from the global SPFN to the latent code of the patch $\mathbf{l}_i$ generated by the local SPFN encoder: $\mathbf{l}_i^{'} = \left[\mathbf{l}_i, \quad \mathbf{l}_o, \quad \mathbf{l}_i^g\right]$. We utilize $\mathbf{l}_i^{'}$ as input to the local SPFN decoder. Our experiments show that providing additional contextual information boosts the performance of the local SPFN as discussed in Section~\ref{sec:ablation}.

\subsection{Segment Merging}
\label{sec:patchMerging}
\vspace{-3pt}

Given a local patch, the local SPFN predicts a segmentation label for each point in the patch where each segment corresponds to a primitive. Our next step is to merge such local per-patch predictions with the predictions of the global SPFN to compute the final segmentation, i.e., primitive decomposition of the high resolution point cloud.

When each local SPFN predicts a maximum of $K_\text{loc}$ segments, we represent the per-point segment label predictions of the $i$-th patch with a probability matrix, $\mathbf{W}_\text{loc}^i \in [0, 1]^{N \times K_\text{loc}}$, which is defined over the \emph{entire} $N$ number of input points:
%
\begin{align}
\vspace{-3pt}
\mathbf{W}_\text{loc}^i = \begin{blockarray}{cccccc}
\begin{block}{(cccccc)}
   & p_{1,1}^i & p_{1,2}^i & \dots & p_{1,K_\text{loc}}^i & & \\
   & p_{2,1}^i & p_{2,2}^i & \dots & p_{2,K_\text{loc}}^i &  & \\
   & \vdots & \vdots &  & \vdots &  & \\
   & p_{N,1}^i & p_{N,2}^i & \dots & p_{N,K_\text{loc}}^i &  & \\
\end{block}
\end{blockarray}
\vspace{-3pt}
\end{align}
%
with $p_{a,b}^i= \mathbb{P}(P_a\in \mathcal{S}_b^i), \quad a\in\{1,..,N\}, \quad b\in\{1,..,K_{\text{loc}}\}$, denoting the probability of point $a$ belonging to segment $b$. Note that a point that does not belong to the patch has zero probability.
 
We represent the prediction results of the global SPFN which predicts a total of $K_\text{glob}$ segments in a similar matrix representation, $\mathbf{W}_\text{glob}$. We then stack each of the segmentation matrices from $M$ patches processed by the local SPFN and the global segmentation matrix:
%
\begin{align}
\vspace{-3pt}
\mathbf{W} = \left[\mathbf{W}_\text{loc}^1 \quad \mathbf{W}_\text{loc}^2 \quad \dots \quad \mathbf{W}_\text{loc}^M \quad \mathbf{W}_\text{glob} \right].
\vspace{-3pt}
\end{align}
%

The goal of the merging step is to compute the one-to-many relationship between the final set of primitives in the input and the individually predicted segmentations. Assuming there are a total of $K_\text{m}$ primitives in the final decomposition, this relationship can be written as a binary matrix $\mathbf{C}  \in \{0, 1\}^{K_\text{m}\times(M\cdot K_\text{loc}+K_\text{glob})}$.

The optimum assignments between the individually predicted segmentations and the final set of primitives need to satisfy certain constraints.
Specifically, each segment should be mapped to exactly one final primitive:
$\mathbf{C}^T \mathds{1}_{K_\text{m}} = \mathds{1}_{M \cdot K_\text{loc}+K_\text{glob}}$.
We further enforce that two segments predicted from the same local patch (or global SPFN) should not be merged under the assumption that the network will avoid over-segmentation:
$\mathbf{C} \mathbf{A} \leq \mathds{1}_{K_\text{m} \times (M+1)}$,
where $\mathbf{A} \in [0, 1]^{(M\cdot K_\text{loc}+K_\text{glob}) \times (M+1)}$ is a matrix indicating the association between segments and patches (or global SPFN).
Finally, we prefer to assign two segments $\mathcal{S}_{k}^i$ and $\mathcal{S}_{l}^j$ predicted from patches $i$ and $j$ to the same final primitive, i.e., merge them, if they have a significant amount of overlap measured as the number of points that belong to both segments. Since $\mathbf{I} = \mathbf{W}^T\mathbf{W}$ represents the intersections between segments as sums of joint probabilities for each point, we find $\mathbf{C}$ by maximizing $\sum_{i,j} \mathbf{I}_{ij} \left(\mathbf{C}^T\mathbf{C}\right)_{ij} = \text{tr}\left(\mathbf{I}\mathbf{C}^T\mathbf{C}\right)$, meaning that we maximize the intersections between the segments assigned to the same final primitive. 

Finally, the final assignment task is formulated as the following binary quadratic programming problem:
%
\begin{equation}
\vspace{-3pt}
\begin{split}
    \mathbf{C}^*   =&  \argmax_{\mathbf{C}} \, \text{tr}\left(\mathbf{I}\mathbf{C}^T\mathbf{C}\right) \\
    \textbf{s.t.} \quad&  \mathbf{C}^T \mathds{1}_{K_\text{m}} = \mathds{1}_{M \cdot K_\text{loc}+K_\text{glob}} \\
    &\mathbf{C} \mathbf{A} \leq \mathds{1}_{K_\text{m} \times (M+1)}.
\end{split}
\end{equation}
%

Finding the optimum $\mathbf{C}^T\mathbf{C}$ instead (a matrix indicating whether two segments are merged or not) also becomes a binary semidefinite programming problem. It typically takes a huge amount of time to solve either the binary quadratic or semidefinite programming. Empirically, we found that a simple heuristic based on Hungarian algorithm~\cite{Kuhn:1955} can provide sufficiently good results while being significantly faster. In $\mathbf{I}$, we find the element that corresponds to the pair of segments with maximum intersection and mark the corresponding indices in $\mathbf{C}^T\mathbf{C}$ as one, i.e., the corresponding pair of segments is merged.  We then zero out all other elements in $\mathbf{I}$ that violate the constraints, thus removing any conflicting segment pairs. These two steps are iterated until no segments can be further merged.
 
Once we recover $\mathbf{C}^*$,  we obtain the final association between the individual points in the input point cloud and the final primitives by computing $\mathbf{W} | {\mathbf{C}^*}^T |^\wedge$, where $\mathbf{X}^\wedge$ is a column-wise $l1$-normalization of $\mathbf{X}$. The output matrix can be considered as including association \emph{scores} from each point to the final primitives, and thus the primitive with the highest score can be picked for each point.
The per-point primitive type is also decided by summing probabilities across the patches including the point and choosing the type with the highest number.
Also, in order to optimize for the corresponding primitive parameters, we need the per-point normals which we estimate as the average of the individual patch-based estimations (we recall that a point can potentially belong to multiple patches resulting in multiple type and normal estimations).

\subsection{Patch Selection Network}
\label{sec:patchSelection}
\vspace{-3pt}

\begin{figure}[h!]
\centering
\newcolumntype{Y}{>{\centering\arraybackslash}X}
\scriptsize
{
\setlength{\tabcolsep}{0.0em}
\renewcommand{\arraystretch}{0.9}
\begin{tabularx}{\linewidth}{CCCCCC}
  GT Segments & $\eta=1\%$ & $\eta=2\%$ & $\eta=3\%$ & $\eta=4\%$ & $\eta=5\%$ \\
  \multicolumn{6}{c}{\includegraphics[width=\linewidth]{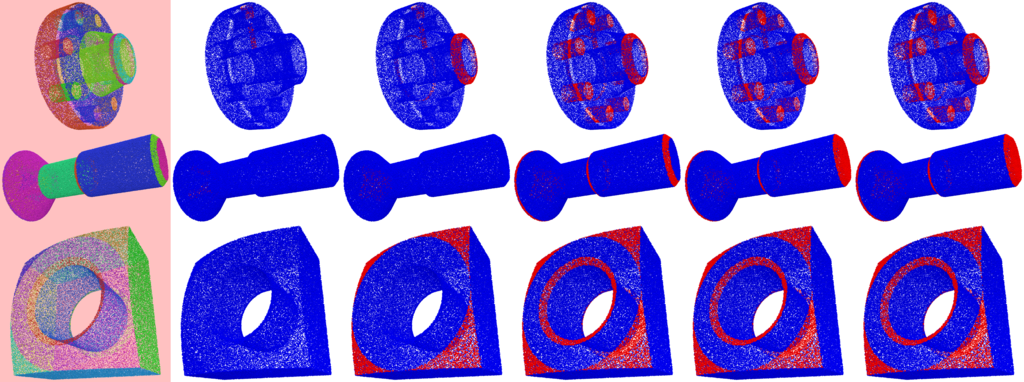}}\\
\end{tabularx}
}
\vspace{-0.5\baselineskip}
\caption{GT Heatmap. Increasing the threshold $\eta$ enlarges the set of small primitives to sample patches from as shown in red.}
\label{fig:gt_heatmap}
\vspace{-1\baselineskip}
\end{figure}

As discussed previously, our method uses a global SPFN module to detect big primitives while relying on the local SPFN to capture small primitives that are likely to be missed in downsampled versions of the input point cloud. In order to ensure the capacity of the local SPFN focuses on small primitive regions, we propose a patch selection strategy both for training and test time. Since at training time we have access to ground truth primitives, we simply sample local patches that belong to small primitives as visualised by the heatmaps in Figure~\ref{fig:gt_heatmap}. We consider a primitive as small if it has less than $\eta\cdot N$ points with $0<\eta<1$ (we experiment with $\eta$ values in the range $[1\%,5\%]$). We randomly sample query points on any such small primitive and generate patches of $n$ points using the $n$-nearest neighbour search for each query point.   

At test time, areas that are likely to contain small primitives are unknown. One naive strategy is to randomly sample local patches to be processed by the local SPFN (see Section~\ref{sec:experiments}). Instead, we design a patch selection network that predicts a heatmap on the input point cloud to predict such areas. The network is trained using the binary cross-entropy loss:
%
\begin{align}
\vspace{-3pt}
    \mathcal{L}_{\text{cross}}=-\sum_{i=1}^n (y_i\log(p_i)+(1-y_i)\log(1-p_i)),
\vspace{-3pt}
\end{align}
%
where $y_i$ is a binary value indicating if point $i$ belongs to a small primitive or not and $p_i$ the estimated probability. Similar to global SPFN, we train the patch selection network on downsampled point clouds. However, we still test the network on low-resolution point clouds which is the resolution at which query points are sampled.

At test time, we binarize the predicted heatmap and generate a pool of points that are likely to belong to small primitives. We randomly sample query points from this pool and generate a local patch. We repeat this process until each point in the pool is covered by at least one patch. Note that the local patches can potentially have overlaps. 
%
%
\section{Experiments}
\label{sec:experiments}
\vspace{-3pt}
%
\begin{figure*}[t!]
\centering
\newcolumntype{Y}{>{\centering\arraybackslash}X}
\scriptsize
{
\setlength{\tabcolsep}{0.0em}
\renewcommand{\arraystretch}{0.9}
\begin{tabularx}{\textwidth}{m{0.08\textwidth}m{0.92\textwidth}}
  \makecell{GT\\Primitives} & \includegraphics[width=0.92\textwidth]{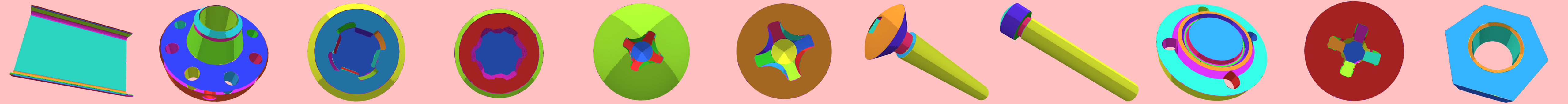} \\
  \makecell{RANSAC\\~\cite{Schnabel:2007}} & \includegraphics[width=0.92\textwidth]{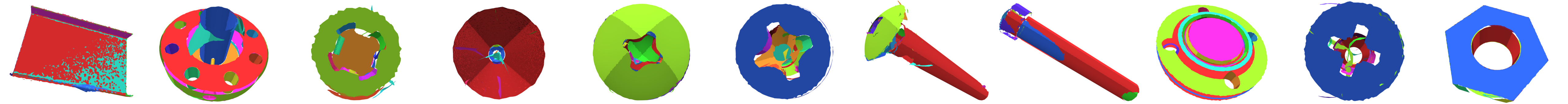} \\
  \makecell{SPFN~\cite{SPFN}} & \includegraphics[width=0.92\textwidth]{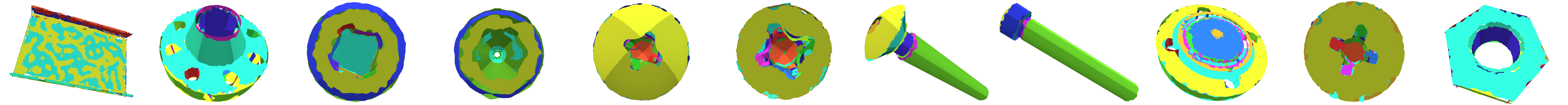} \\
  \makecell{CPFN - 5\%} & \includegraphics[width=0.92\textwidth]{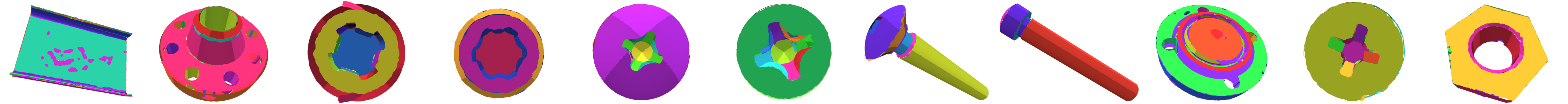} \\
  \midrule
  \makecell{GT Heatmap\\$\eta=5\%$} & \includegraphics[width=0.92\textwidth]{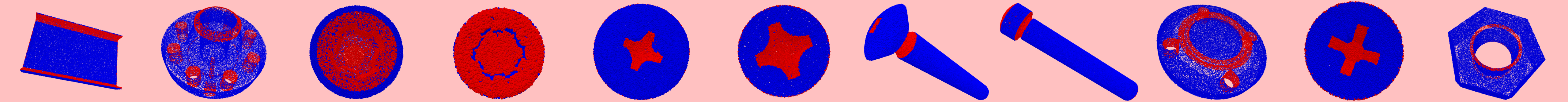} \\
  \makecell{Predicted\\Heatmap\\$\eta=5\%$} & \includegraphics[width=0.92\textwidth]{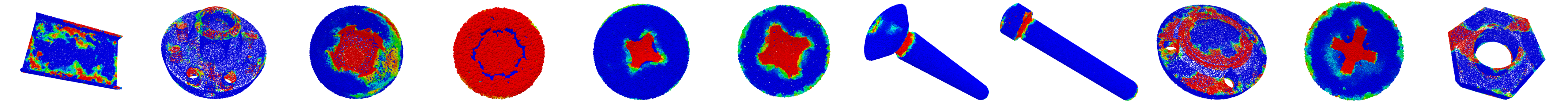} \\
\end{tabularx}
}
\vspace{-0.5\baselineskip}
\caption{Primitive fitting results for RANSAC, plain SPFN and our CPFN networks. RANSAC and SPFN directly operate on the global object failing on the small primitives. Our CPFN pipeline estimates the heatmaps corresponding to small primitives at different scales and learns a better primitive decomposition on local patches sampled from such regions improving the detection of small primitives. Predicted heatmaps are displayed with the \textit{Jet} color map going from blue to green to red. See the supplementary for more results.}
\label{fig:results}
\vspace{-1\baselineskip}
\end{figure*}
%

\subsection{Dataset}
\label{subsec:dataset}
\vspace{-3pt}

We evaluate our proposed method on CAD models from American National Standards Institute (ANSI)~\cite{ANSI} mechanical components provided by TraceParts~\cite{TraceParts}. SPFN~\cite{SPFN} provides the pre-processed point clouds at two resolutions of $8$k and $128$k points. In our experiments, we use these high-resolution point clouds already provided if not stated otherwise, i.e., $N=128$k.

The dataset consists of $504$ categories, and we use the same training/test splits introduced in~\cite{SPFN}, providing $13\,831/3\,366$ models for each, respectively.
Similar to the low-resolution version of the data, the high-resolution point clouds are preprocessed in a way to merge adjacent primitives sharing the same parameters and discard extremely tiny primitives, which have an area less than $0.5\%$ of the entire object.
(Note that the low-resolution version discarded primitives with less than $2\%$ of the area, and thus losing more small primitives than the high-resolution version.)
The point clouds are normalized to the unit sphere and also randomly perturbed with uniform noise $[-5\text{e-}3,5\text{e-}3]$ along the ground truth normal direction.
The maximum number of primitives in a given object is 28. Thus, we set $K_{\text{glob}}=28$. 

The input to our network is the noisy point cloud with the random noise, denoted by $\mathbf{P}$.

The dataset also provides $512$ points sampled on each of the ground-truth primitive surfaces (denoted by $\{\mathbf{S}_k\}$), which we use to compute the residual errors of our predictions similar to SPFN.

To train our global SPFN, we downsample the high-resolution point clouds to have $n=8$k points using Furthest Point Sampling (FPS).
This ensures that we also obtain points on the very tiny primitives preserved in the high-resolution data but discarded in the low-resolution version.

To train our local SPFN, we sample patches located in areas that contain small primitives (with less than $\eta\cdot N$ points) based on the ground truth primitive decomposition. Specifically, we first select a pool of points in the low-resolution point cloud that belong to any of the small primitives. Then, patches of $n$ points are extracted from the high-resolution point cloud using $n$-nearest neighbors by randomly sampling query points from our pool. The sampling process stops when all points in the pool are covered by a patch. Each patch is then centered at the origin and scaled to the unit sphere. Per-point normals and primitive parameters are modified to take into account these transformations. We set $K_{\text{loc}}=21$ which is the maximum number of primitives observed on a single patch.

\subsection{Evaluation metrics}
\label{subsec:metrics}
\vspace{-3pt}

We use the same seven evaluation metrics reported in~\cite{SPFN}: (i) segmentation mean intersection over union (mIoU) in \%, (ii) mean primitive type accuracy in \%, (iii) mean point normal difference in degrees, $^{\circ}$, (iv) mean primitive axis difference in degrees, $^{\circ}$, (v) mean/std. $\{\mathbf{S}_k\}$ residual, (vi) $\{\mathbf{S}_k\}$ coverage in \% and (vii) $\mathbf{P}$ coverage in \%. 

(i), (ii) and (iv) are computed for all primitives and are then averaged for each point cloud. As mentioned previously, the correspondence between the ground truth and predicted primitives is computed with the Hungarian matching algorithm so that the mIoU is maximized. In these metrics, we weight each primitive the same independently from their scales. (iii) is computed for each point and then averaged for all points within the point cloud.

(v) assesses the fitting error of each predicted primitive $\mathbf{S}_k$ using the $512$ points pre-sampled on the assigned ground truth primitive surfaces.  (vi) computes for each primitive the proportion of the pre-sampled points that are closer to the predicted primitive surface than an $\epsilon$ distance. Both (v) and (vi) are then averaged for each point cloud.

(vii) reports the proportion of points $\mathbf{P}$ that are closer to any predicted primitive surface than an $\epsilon$ distance.

We finally report the average of all of these metrics over all the point clouds in the test set.

%
\begin{table*}[t!]
\centering
\caption{
Quantitative comparisons across our CPFN, other baseline methods, and ablation cases. $\mathbf{l}_i$, $\mathbf{l}_o$, and $\mathbf{l}_i^g$ indicate patch features from Local SPFN, object features from global SPFN, patch features from global SPFN, respectively (see Section~\ref{sec:patchPrediction}), and GP and PN stand for Global SPFN and Patch Selection Network (Section~\ref{sec:patchSelection}). If PN is not used, patches are sampled randomly to cover the whole object.
}.
\newcolumntype{Y}{>{\centering\arraybackslash}X}
\footnotesize{
{
\setlength{\tabcolsep}{0.2em}
\renewcommand{\arraystretch}{0.9}
\begin{tabularx}{\textwidth}{>{\centering}m{0.3cm}|>{\centering}m{0.5cm}|>{\centering}m{0.5cm}|>{\centering}m{0.5cm}|>{\centering}m{0.5cm}|>{\centering}m{0.5cm}|Y|Y|Y|Y|>{\centering}m{1.8cm}|Y|Y|Y|Y}
  \toprule
    \multirow{2}{*}{Id} &
    \multicolumn{5}{c|}{\multirow{2}{*}{Method}} &
    \multirow{2}{*}{{\makecell{Seg. (Mean\\IoU) (\%) $\uparrow$}}} &
    \multirow{2}{*}{{\makecell{Primitive\\Type (\%) $\uparrow$}}} &
    \multirow{2}{*}{{\makecell{Point\\Normal ($^{\circ}$) $\downarrow$}}} &
    \multirow{2}{*}{{\makecell{Primitive\\Axis ($^{\circ}$) $\downarrow$}}} &
    \multirow{2}{*}{{\makecell{$\{\mathbf{S}_k\}$ Residual\\Mean $\pm$ Std. $\downarrow$}}} &
    \multicolumn{2}{c|}{$\{\mathbf{S}_k\}$ Coverage (\%) $\uparrow$} &
    \multicolumn{2}{c}{$\mathbf{P}$ Coverage (\%) $\uparrow$} \\
  \cline{12-15}
     & \multicolumn{5}{c|}{} & & & & & &
     $\epsilon = 0.01$ & $\epsilon = 0.02$ & $\epsilon = 0.01$ & $\epsilon = 0.02$ \\
  \midrule
    1 & \multicolumn{5}{c|}{RANSAC~\cite{Schnabel:2007}} & 55.01 & 59.14 & 11.16 & 3.24 & 0.073 $\pm$ 0.039 & 61.88 & 67.66 & 78.93 & 86.64  \\
    2 & \multicolumn{5}{c|}{SPFN~\cite{SPFN}} & 66.29 & 89.50 & 10.59 & 1.25 & \textbf{0.020} $\pm$ \textbf{0.010} & 72.94 & 82.31 & 88.69 & \textbf{94.57} \\
    3 & \multicolumn{5}{c|}{CPFN \scriptsize{($\eta=1\%$)}} & 70.11 & 93.49 & 9.37 & 1.37 & 0.026 $\pm$ 0.013 & 72.58 & 80.88 & 87.60 & 93.76 \\
    4 & \multicolumn{5}{c|}{CPFN \scriptsize{($\eta=2\%$)}} & 75.85 & 95.55 & 7.79 & \textbf{1.17} & 0.034 $\pm$ 0.017 & 72.94 & 79.58 & 87.36 & 92.69\\
    5 & \multicolumn{5}{c|}{CPFN \scriptsize{($\eta=3\%$)}} & 78.45 & 96.22 & 6.87 & 1.39 & 0.033 $\pm$ 0.017 & 75.46 & 81.23 & 88.84 & 93.11 \\
    6 & \multicolumn{5}{c|}{CPFN \scriptsize{($\eta=4\%$)}} & 79.09 & 96.37 & 6.63 & 1.48 & 0.032 $\pm$ 0.016 & 76.63 & 82.18 & \textbf{89.19} & 93.41 \\
    7 & \multicolumn{5}{c|}{CPFN \scriptsize{($\eta=5\%$)}} & \textbf{79.64} & \textbf{96.45} & \textbf{6.48} & 1.44 & 0.030 $\pm$ 0.015 & \textbf{76.64} & \textbf{82.54} & 88.73 & 93.12 \\
  \midrule
    & $\mathbf{l}_i$ & $\mathbf{l}_o$ & $\mathbf{l}_i^g$ & GS & PN & \multicolumn{9}{c}{Ablation Study (CPFN, $\eta=5\%$)}\\
  \midrule
    8 & \checkmark &  &  & \checkmark & \checkmark & 77.70 & 95.12 & 6.68 & 3.07 & 0.053 $\pm$ 0.029 & 64.55 & 70.60 & 85.21 & 90.53 \\
    9 & \checkmark & \checkmark &  & \checkmark & \checkmark & 78.30 & 95.96 & 6.61 & 1.70 & 0.033 $\pm$ 0.016 & 74.43 & 80.92 & 87.99 & 92.55\\
    10 & \checkmark &  & \checkmark & \checkmark & \checkmark &  77.95 & 95.55 & 6.65 & 2.51 & 0.052 $\pm$ 0.026 & 64.72 & 70.92 & 86.00 & 91.37\\
    11 & \checkmark & \checkmark & \checkmark &  &  & 74.13  & 91.97 & \textbf{5.15} & 1.85  & 0.063 $\pm$ 0.022 & 71.09 & 75.73 & 87.15 & 91.61 \\
    12 & \checkmark & \checkmark & \checkmark & \checkmark &  & 78.44 & 95.47 & 5.28 & 1.67 & 0.052 $\pm$ 0.026 & 50.62 & 62.36 & 74.42 & 83.65 \\
    13 & \checkmark & \checkmark & \checkmark & \checkmark & \checkmark & \textbf{79.64} & \textbf{96.45} & 6.48 & \textbf{1.44} & \textbf{0.030} $\pm$ \textbf{0.015} & \textbf{76.64} & \textbf{82.54} & \textbf{88.73} & \textbf{93.12} \\
  \midrule
    14 & \multicolumn{5}{c|}{CPFN w/ GT Heatmap} & 80.94 & 96.74 & 6.83 & 1.45 & 0.028 $\pm$ 0.014 & 79.45 & 84.94 & 90.63 & 94.55 \\
  \bottomrule
\end{tabularx}
}
}
\vspace{-0.5\baselineskip}
\label{tab:results_traceparts}
\end{table*}
%
\begin{table}[ht!]
\footnotesize{
\setlength{\tabcolsep}{2.0pt}
  \centering
  \caption{
  The accuracy of CPFN, SPFN, and RANSAC in detecting primitives with varying scales in terms of mIoU ($\%$). GP and PN stand for Global SPFN and Patch Selection Network (Section~\ref{sec:patchSelection}). Each scale bucket contains roughly the same number of primitives.
  }
    \begin{tabularx}{\linewidth}{>{\centering}m{1.0cm}|>{\centering}m{1.0cm}|CCCCC}
    \toprule
    \multicolumn{2}{c|}{Scale} & $\sim$1\% & 1\%$\sim$2\% & 2\%$\sim$4\% & 4\%$\sim$12\% & 12\%$\sim$ \\
    \midrule
    \multicolumn{2}{c|}{RANSAC~\cite{Schnabel:2007}} & 34.68 & 40.38 & 56.78 & 70.63 & 69.50 \\
    \multicolumn{2}{c|}{SPFN~\cite{SPFN}}  & 44.25 & 55.53 & 70.12 & 74.29 & 79.75 \\
    \multicolumn{2}{c|}{CPFN \scriptsize{($\eta=1\%$)}}& 56.21 & 61.93 & 71.91 & 76.04 & 80.02 \\
    \multicolumn{2}{c|}{CPFN \scriptsize{($\eta=2\%$)}}& 63.50 & 73.31 & 78.06 & 79.31 & 81.93 \\
    \multicolumn{2}{c|}{CPFN \scriptsize{($\eta=3\%$)}}& 64.89 & 76.19 & 82.85 & 81.62 & 83.54 \\
    \multicolumn{2}{c|}{CPFN \scriptsize{($\eta=4\%$)}}& 65.23 & 76.37 & 83.71 & 82.85 & 83.68 \\
    \multicolumn{2}{c|}{CPFN \scriptsize{($\eta=5\%$)}}& \textbf{65.74} & \textbf{77.31} & \textbf{84.19} & \textbf{83.55} & \textbf{83.95} \\
    \midrule
    GS & PN & \multicolumn{5}{c}{Ablation Study (CPFN, $\eta=5\%$)}\\
  \midrule
     & & 52.38 & 66.81 & 64.36 & 79.24 & \textbf{88.24}\\
    \checkmark & & 60.10 & 75.32 & 82.71 & \textbf{83.83} & 85.38 \\
    \checkmark & \checkmark & \textbf{65.74} & \textbf{77.31} & \textbf{84.19} & 83.55 & 83.95 \\
    \bottomrule
    \end{tabularx}
  \label{tab:miou_primitives_scale}
}
\vspace{-1.5\baselineskip}
\end{table}
%

\subsection{Results}
\label{subsec:results}
\vspace{-3pt}

\vspace{-3pt}
\paragraph{Comparison with Global SPFN} As a baseline, we first show the case when using only the global branch of our method. At test time, as the memory limitation due to PointNet++ is lifted, we evaluate the performance of the global SPFN on high-resolution point clouds. As shown in Table~\ref{tab:results_traceparts}, the original SPFN (row 2) achieves an mIoU of $66.29\%$ on the original test set~\footnote{Note that we use a high-resolution version of the dataset that also includes more small primitives. Thus, the numbers reported here are different from the ones in~\cite{SPFN}.}. As shown numerically in Table~\ref{tab:miou_primitives_scale} and visually in Figure~\ref{fig:results}, the performance highly depends on the size of the primitives and significantly drops when only small primitives are considered. Specifically, for primitives that have a scale of less than $1\%$, i.e., primitives that contain less than $\eta \cdot N$ points, $\eta\leq 1\%$, mIoU drops to $44.25\%$. Similarly, for $1\%\leq\eta\leq 2\%$, mIoU is $55.53\%$.

\vspace{-3pt}
\paragraph{CPFN} We train the local, patch-supervised primitive fitting network (local SPFN) with patches sampled on ground truth small primitives and test on patches identified by our patch selection network. During training, we consider $5$ different scales, $\eta\in\{1\%,2\%,3\%,4\%,5\%\}$, to identify small primitives and hence train $5$ different versions of our local SPFN. In Table~\ref{tab:results_traceparts}, we report the accuracy of the complete pipeline, i.e., CPFN, using each of the local SPFN versions SPFN (rows 3-7). We further report the accuracy of each local SPFN version at the patch level in Table~\ref{tab:results_patches}.

As shown by the quantitative numbers, we improve the accuracy of the original SPFN with respect to almost all the metrics. Specifically, we achieve a significant improvement in terms of mIoU ($+13.35\%$),  type accuracy ($+6.95\%$), point normal accuracy (decreasing difference by $-38.81\%$), and $\{\mathbf{S}_k\}$ coverage (+3.70\% at $\epsilon=0.01$) for $\eta=5\%$.

We note that the patches selected by our selection network do not solely contain small primitives but also a portion of larger primitives. Thus, CPFN has a positive impact both on small and larger primitives, especially improving the predicted normals in high curvature areas between the primitives. Also, by improving the segmentation on the smaller primitives (Figure~\ref{fig:results}), CPFN achieves a substantial boost in mIoU performance as the metric is independent to the size of the primitive. Differently, small primitives have a lower weight in the computation of $\mathbf{P}$ coverage thus limiting the improvement gain on this metric.

Increasing the scale of the primitives to be used for sampling training patches of local SPFN improves the overall performance. As shown in Tables~\ref{tab:results_traceparts} and~\ref{tab:miou_primitives_scale}, CPFN with $\eta=5\%$ local SPFN generally outperforms the other versions. However, as we discuss in the ablation, training local SPFN on all patches without considering any threshold $\eta$ results in lower accuracy. Our patch selection strategy ensures the capacity of local SPFN is utilized to improve the detection of small primitives, the main bottleneck of SPFN. 

%
\begin{table*}[ht!]
\centering
\caption{
Performance of the local SPFNs trained with patches sampled from primitives with varying scales, $\eta\in\{1\%,2\%,3\%,4\%,5\%\}$, tested at patch level.
}
\newcolumntype{Y}{>{\centering\arraybackslash}X}
\footnotesize{
\setlength{\tabcolsep}{0.2em}
\renewcommand{\arraystretch}{0.9}
\begin{tabularx}{\textwidth}{m{3.1cm}|Y|Y|Y|Y|>{\centering}m{1.8cm}|Y|Y|Y|Y}
  \toprule
    \multirow{2}{*}{Method} &
    \multirow{2}{*}{{\makecell{Seg. (Mean\\IoU) (\%) $\uparrow$}}} &
    \multirow{2}{*}{{\makecell{Primitive\\Type (\%) $\uparrow$}}} &
    \multirow{2}{*}{{\makecell{Point\\Normal ($^{\circ}$) $\downarrow$}}} &
    \multirow{2}{*}{{\makecell{Primitive\\Axis ($^{\circ}$) $\downarrow$}}} &
    \multirow{2}{*}{{\makecell{$\{\mathbf{S}_k\}$ Residual\\Mean $\pm$ Std. $\downarrow$}}} &
    \multicolumn{2}{c|}{$\{\mathbf{S}_k\}$ Coverage (\%) $\uparrow$} &
    \multicolumn{2}{c}{$\mathbf{P}$ Coverage (\%) $\uparrow$} \\
  \cline{7-10}
     & & & & & &
     $\epsilon = 0.01$ & $\epsilon = 0.02$ & $\epsilon = 0.01$ & $\epsilon = 0.02$ \\
  \midrule
  Local SPFN \scriptsize{($\eta=1\%$)}&
  74.39 & 93.12 & 6.92 & 1.87 & 0.097 $\pm$ 0.063 & 56.96 & 65.26 & 87.70 & 93.20 \\
  Local SPFN \scriptsize{($\eta=2\%$)}&
  80.91 & 95.77 & 6.12 & 1.37 & \textbf{0.079} $\pm$ \textbf{0.052} & \textbf{64.87} & \textbf{72.06} & 90.65 & 94.54  \\
  Local SPFN \scriptsize{($\eta=3\%$)}&
  81.53 & 96.23 & 6.06 & 1.33 & 0.087 $\pm$ 0.057 & 63.50 & 70.48 & 90.90 & 94.78  \\
  Local SPFN \scriptsize{($\eta=4\%$)}&
  81.42 & \textbf{96.32} & \textbf{6.01} & 1.35 & 0.086 $\pm$ 0.058 & 63.72 & 70.70 & \textbf{90.99} & \textbf{94.87}  \\
  Local SPFN \scriptsize{($\eta=5\%$)}&
  \textbf{81.60} & 96.07 & 6.04 & \textbf{1.29} & 0.089 $\pm$ 0.059 & 62.77 & 69.86 & 90.53 & 94.58  \\
  \bottomrule
\end{tabularx}
}
\vspace{-\baselineskip}
\label{tab:results_patches}
\end{table*}

%

\vspace{-7pt}
\paragraph{Comparison with RANSAC}
We compare our method with the commonly used RANSAC-based approach for primitive fitting. Specifically, we use the efficient RANSAC~\cite{Schnabel:2007} implementation provided by CGAL~\cite{CgalRANSAC}. We use the default parameters except for the maximum point-to-surface distance for which we choose a value equal to twice the noise level $2\cdot\nu=0.01$.

As shown in Table~\ref{tab:results_traceparts}, RANSAC (row 1) under-performs on all metrics compared to both SPFN and CPFN. It is specifically prone to either under- or over-segmentation (see Figure~\ref{fig:results}) due to noise resulting in a segmentation accuracy as low as $55.01\%$ mIoU. Due to ambiguity in primitive types, it also has a low type prediction accuracy of $59.14\%$. Table~\ref{tab:miou_primitives_scale} shows that, similar to SPFN, the performance of RANSAC significantly drops for smaller primitives ($34.68\%$ and $40.38\%$ mIoU for $\eta\leq 1\%$ and $1\%\leq\eta\leq 2\%$ respectively). The poor segmentation performance has a direct impact on the prediction of primitive axes, residual loss, and coverage.

\vspace{-7pt}
\paragraph{Effect of Point Cloud Resolution}
To assess the performance of our method on different resolutions, we experimented with two additional resolutions, 16k and 64k, as shown in Table~\ref{tab:results_resolution_impact}.
The comparison between the results of SPFN and our CPFN shows that \textit{higher resolutions} benefit more from our two-level prediction architecture.

%
\vspace{-0.5\baselineskip}
\begin{table}[ht!]
\footnotesize{
\setlength{\tabcolsep}{2.0pt}
  \centering
  \caption{Quantitative comparison of SPFN~\cite{SPFN} and CPFN with different resolutions of the point clouds.}
    \begin{tabularx}{\linewidth}{c|CC|CC|CC}
    \toprule
    Res. & \multicolumn{2}{c|}{16k} & \multicolumn{2}{c|}{64k} & \multicolumn{2}{c}{128k} \\ 
    \midrule
    Method &
    SPFN & CPFN &
    SPFN & CPFN &
    SPFN & CPFN \\
    \midrule
    Mean IoU (\%) & 65.86 & 74.77  & 66.25 & 78.29 & 66.30 & \textbf{79.64} \\
    \bottomrule
    \end{tabularx}
  \label{tab:results_resolution_impact}
}
\vspace{-1.3\baselineskip}
\end{table}
%

\begin{figure}[h!]
\centering
\vspace{-\baselineskip}
\includegraphics[width=\linewidth]{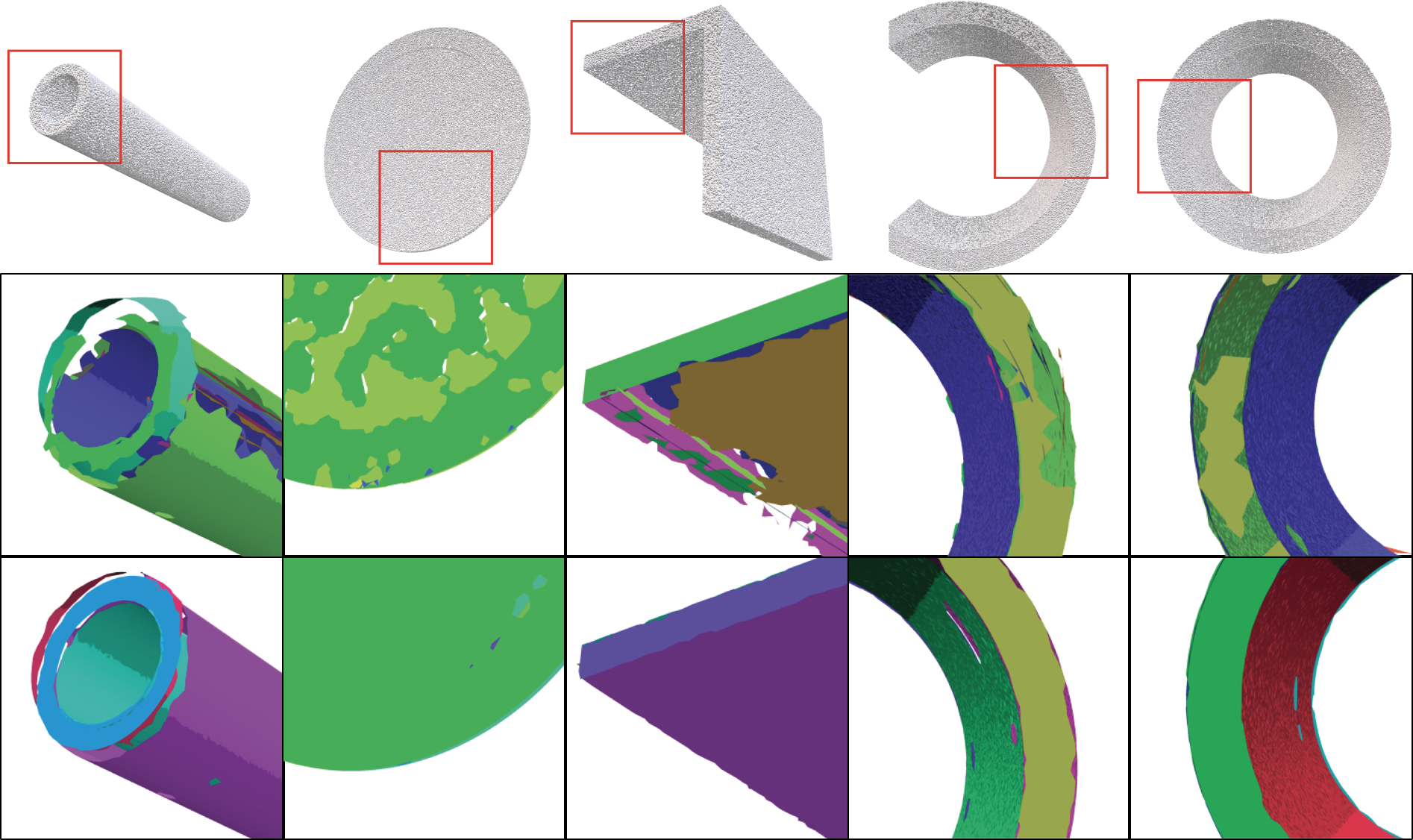}
\caption{
Generalization test results with ABC dataset. From top to bottom, inputs, SPFN~\cite{SPFN} results, and CPFN results (in red box areas). Both networks are trained with TraceParts dataset.
}
\label{fig:generalization}
\vspace{-1\baselineskip}
\end{figure}

\paragraph{Generalization}
\label{sec:abc_results}
To evaluate the generalization power of CPFN, we test how it performs on shapes that are acquired from different datasets than it was trained on. As shown in Figure~\ref{fig:generalization}, while the use of local patches improves the generalization capability of CPFN, as the shapes become significantly different than those seen during training, the performance degrades.

%
%

\subsection{Ablation Study}
\label{sec:ablation}
\vspace{-3pt}

We motivate the different design choices of our method with a detailed ablation study. We first assess the effect of using additional contextual information as input to our local SPFN. Specifically, we provide both object level global features $\mathbf{l}_o$ and patch level local features $\mathbf{l}_i^g$, both extracted by the global SPFN, as additional input to the local features extracted from the local SPFN $\mathbf{l}_i$ (See Section~\ref{sec:patchPrediction}). As shown in Table~\ref{tab:results_traceparts}, the use of global features $\mathbf{l}_o$ has a positive impact on the patch segmentation accuracy ($+0.60\%$ mIoU, see row 9). More importantly, global features help to reduce the error in predicting primitive parameters that depend on global context such as primitive axis ($44.63\%$ improvement). This is further reflected by an improved $\{\mathbf{S}_k\}$ residual by $37.74\%$ and a boost in $\{\mathbf{S}_k\}$ coverage ($+9.88\%$ at $\epsilon=0.01$). The  local  features from the global SPFN $\mathbf{l}_i^g$ have a similar positive impact (see row 10). When both features are used in combination, we see a more significant improvement: an increase of $+1.94\%$ in segmentation mIoU, an increase of $+1.33\%$ in type prediction accuracy, a reduction of $-53.09\%$ in prediction of primitive axis error, and an increase of $+12.09\%$ in $\{\mathbf{S}_k\}$ coverage at $\epsilon=0.01$.

We also test our CPFN without using the global SPFN predictions in the merging process (row 11).
To cover the entire shape, here we train our local SPFN without the patch selection network (Section~\ref{sec:patchSelection}) but randomly sample patches over the entire input point cloud.
Compared to our full CPFN pipeline (with $\eta=5\%$), the overall mIoU drops by $5.51\%$ with this alternative pipeline.
We further evaluate CPFN without using the patch selection network (i.e., process all sampled patches by the local SPFN) but still using the global SPFN in the merging step (row 12). The segmentation mIoU slightly decreases compared to the full CPFN baseline ($-1.09\%$).
Lastly, we also evaluate the impact of the possible errors introduced by the patch selection network (row 14). With ground truth patch selection provided at test time, we see only a marginal improvement (an improvement of $+1.30\%$ in segmentation mIoU) showing that our method is not very sensitive to patch selection errors.
\section{Conclusion}
\vspace{-3pt}

We presented CPFN, a cascaded primitive fitting network that focuses on fitting primitives to high-resolution point clouds obtained by scanning. Our approach consists of a cascade of global fitting network that operates on a downsampled version of the input point cloud as well as a local fitting network that processes local patches on the high resolution point cloud. We present a novel merging formulation that ensembles global and local predictions that outperform state-of-the-art fitting results, especially in regions of fine-scale details.
In future work, we would like to explore developing a fully end-to-end trainable pipeline for all of the patch selection, patch-based prediction, and merging steps.

\vspace{-3pt}
{\footnotesize
\paragraph{Acknowledgements}
The ANSI mechanical component CAD models are originally provided by TraceParts and curated by Li~\etal~\cite{SPFN}.
M.~Sung acknowledges the support by the National Research Foundation~(NRF) grant funded by the Korea government (MSIT) (2021R1F1A1045604).
}

{\small
\bibliographystyle{ICCV/ieee_fullname}
\bibliography{egbib}

\begin{thebibliography}{10}\itemsep=-1pt

\bibitem{ANSI}
American {N}ational {S}tandards {I}nstitute ({ANSI}).

\bibitem{Binford:1971}
Thomas~O. Binford.
\newblock Visual perception by computer.
\newblock In {\em {IEEE} Conference on Systems and Control}, 1971.

\bibitem{Chen:2020}
J. {Chen}, B. {Lei}, Q. {Song}, H. {Ying}, D.~Z. {Chen}, and J. {Wu}.
\newblock A hierarchical graph network for 3d object detection on point clouds.
\newblock In {\em CVPR}, 2020.

\bibitem{Chen:2019}
S. {Chen}, S. {Niu}, T. {Lan}, and B. {Liu}.
\newblock Pct: Large-scale 3d point cloud representations via graph inception
  networks with applications to autonomous driving.
\newblock In {\em ICIP}, 2019.

\bibitem{BSP-Net}
Z. {Chen}, A. {Tagliasacchi}, and H. {Zhang}.
\newblock {BSP-Net}: Generating compact meshes via binary space partitioning.
\newblock In {\em CVPR}, 2020.

\bibitem{Deng:2020}
B. {Deng}, K. {Genova}, S. {Yazdani}, S. {Bouaziz}, G. {Hinton}, and A.
  {Tagliasacchi}.
\newblock {CvxNet}: Learnable convex decomposition.
\newblock In {\em CVPR}, 2020.

\bibitem{Fischler:1981}
Martin~A. Fischler and Robert~C. Bolles.
\newblock Random sample consensus: A paradigm for model fitting with
  applications to image analysis and automated cartography.
\newblock {\em Communications of the ACM}, 1981.

\bibitem{shapehandles}
Matheus Gadelha, Giorgio Gori, Duygu Ceylan, Radomir Mech, Nathan Carr, Tamy
  Boubekeur, Rui Wang, and Subhransu Maji.
\newblock Learning generative models of shape handles.
\newblock In {\em CVPR}, 2020.

\bibitem{Genova:2020}
K. {Genova}, F. {Cole}, A. {Sud}, A. {Sarna}, and T. {Funkhouser}.
\newblock Local deep implicit functions for 3d shape.
\newblock In {\em CVPR}, 2020.

\bibitem{Genova:2019}
K. {Genova}, F. {Cole}, D. {Vlasic}, A. {Sarna}, W. {Freeman}, and T.
  {Funkhouser}.
\newblock Learning shape templates with structured implicit functions.
\newblock In {\em ICCV}, 2020.

\bibitem{Graham:2018}
Benjamin Graham, Martin Engelcke, and Laurens van~der Maaten.
\newblock {3D} semantic segmentation with submanifold sparse convolutional
  networks.
\newblock In {\em CVPR}, 2018.

\bibitem{gurobi}
LLC Gurobi~Optimization.
\newblock Gurobi optimizer reference manual, 2020.

\bibitem{Han:2020}
L. {Han}, T. {Zheng}, L. {Xu}, and L. {Fang}.
\newblock {OccuSeg}: Occupancy-aware 3d instance segmentation.
\newblock In {\em CVPR}, 2020.

\bibitem{Hu:2020}
Q. {Hu}, B. {Yang}, L. {Xie}, S. {Rosa}, Y. {Guo}, Z. {Wang}, N. {Trigoni}, and
  A. {Markham}.
\newblock {RandLA-Net}: Efficient semantic segmentation of large-scale point
  clouds.
\newblock In {\em CVPR}, 2020.

\bibitem{Jiang:2020}
L. {Jiang}, H. {Zhao}, S. {Shi}, S. {Liu}, C.~W. {Fu}, and J. {Jia}.
\newblock {PointGroup}: Dual-set point grouping for 3d instance segmentation.
\newblock In {\em CVPR}, 2020.

\bibitem{Kaiser:2018}
Adrien Kaiser, Jose Alonso~Ybanez Zepeda, and Tamy Boubekeur.
\newblock A survey of simple geometric primitives detection methods for
  captured {3D} data.
\newblock {\em Computer Graphics Forum}, 2018.

\bibitem{UCSG-Net}
Kacper Kania, Maciej Zi{\k{e}}ba, and Tomasz Kajdanowicz.
\newblock {UCSG-Net} -- unsupervised discovering of constructive solid geometry
  tree, 2020.
\newblock arXiv:2006.09102.

\bibitem{adam}
Diederik~P. Kingma and Jimmy Ba.
\newblock Adam: {A} method for stochastic optimization.
\newblock In Yoshua Bengio and Yann LeCun, editors, {\em 3rd International
  Conference on Learning Representations, {ICLR} 2015, San Diego, CA, USA, May
  7-9, 2015, Conference Track Proceedings}, 2015.

\bibitem{Kuhn:1955}
H.~W. Kuhn.
\newblock The hungarian method for the assignment problem.
\newblock {\em Naval Research Logistics Quarterly}, 1955.

\bibitem{CSG}
David~H. Laidlaw, W.~Benjamin Trumbore, and John~F. Hughes.
\newblock Constructive solid geometry for polyhedral objects.
\newblock In {\em SIGGRAPH}, 1986.

\bibitem{Landrieu:2018}
L. {Landrieu} and M. {Simonovsky}.
\newblock Large-scale point cloud semantic segmentation with superpoint graphs.
\newblock In {\em CVPR}, 2018.

\bibitem{SPFN}
Lingxiao Li, Minhyuk Sung, Anastasia Dubrovina, Li Yi, and Leonidas Guibas.
\newblock Supervised fitting of geometric primitives to {3D} point clouds.
\newblock In {\em CVPR}, 2019.

\bibitem{li_globFit_sigg11}
Yangyan Li, Xiaokun Wu, Yiorgos Chrysanthou, Andrei Sharf, Daniel Cohen-Or, and
  Niloy~J. Mitra.
\newblock {GlobFit}: Consistently fitting primitives by discovering global
  relations.
\newblock In {\em SIGGRAPH}, 2011.

\bibitem{Lin:2020}
Cheng Lin, Tingxiang Fan, Wenping Wang, and Matthias Nießner.
\newblock Modeling {3D} shapes by reinforcement learning.
\newblock In {\em ECCV}, 2020.

\bibitem{Marr:1978}
D. Marr and H.~K. Nishihara.
\newblock Representation and recognition of the spatial organization of
  three-dimensional shapes.
\newblock {\em Proceedings of the Royal Society of London B: Biological
  Sciences}, 1978.

\bibitem{CgalRANSAC}
Sven Oesau, Yannick Verdie, Cl{\'e}ment Jamin, Pierre Alliez, Florent Lafarge,
  and Simon Giraudot.
\newblock Point set shape detection.
\newblock In {\em {CGAL} User and Reference Manual}. {CGAL Editorial Board},
  {4.13} edition, 2018.

\bibitem{Paschalidou:2019}
D. {Paschalidou}, A.~O. {Ulusoy}, and A. {Geiger}.
\newblock Superquadrics revisited: Learning 3d shape parsing beyond cuboids.
\newblock In {\em CVPR}, 2019.

\bibitem{PointNet}
Charles~Ruizhongtai Qi, Hao Su, Kaichun Mo, and Leonidas~J. Guibas.
\newblock {PointNet}: Deep learning on point sets for {3D} classification and
  segmentation.
\newblock In {\em CVPR}, 2017.

\bibitem{PointNet++}
Charles~Ruizhongtai Qi, Ly Yi, Hao Su, and Leonidas~J. Guibas.
\newblock {PointNet++}: Deep hierarchical feature learning on point sets in a
  metric space.
\newblock In {\em NIPS}, 2017.

\bibitem{Rethage:2018}
Dario Rethage, Johanna Wald, J{\"{u}}rgen Sturm, Nassir Navab, and Federico
  Tombari.
\newblock Fully-convolutional point networks for large-scale point clouds.
\newblock In {\em ECCV}, 2018.

\bibitem{OctNet}
G. {Riegler}, A.~O. {Ulusoy}, and A. {Geiger}.
\newblock {OctNet}: Learning deep 3d representations at high resolutions.
\newblock In {\em CVPR}, 2017.

\bibitem{TraceParts}
TraceParts S.A.S.
\newblock Traceparts.

\bibitem{Schnabel:2007}
Ruwen Schnabel, Roland Wahl, and Reinhard Klein.
\newblock Efficient {RANSAC} for point-cloud shape detection.
\newblock {\em Computer Graphics Forum}, 2007.

\bibitem{Sharma:2018}
Gopal Sharma, Rishabh Goyal, Difan Liu, Evangelos Kalogerakis, and Subhransu
  Maji.
\newblock {CSGNet}: Neural shape parser for constructive solid geometry.
\newblock In {\em CVPR}, 2018.

\bibitem{Sharma:2020}
Gopal Sharma, Difan Liu, Evangelos Kalogerakis, Subhransu Maji, Siddhartha
  Chaudhuri, and Radom\'{i}r M\v{e}ch.
\newblock {ParSeNet}: A parametric surface fitting network for 3d point clouds.
\newblock In {\em ECCV}, 2020.

\bibitem{Smirnov:2020}
D. {Smirnov}, M. {Fisher}, V.~G. {Kim}, R. {Zhang}, and J. {Solomon}.
\newblock Deep parametric shape predictions using distance fields.
\newblock In {\em CVPR}, 2020.

\bibitem{Sun:2019}
Chunyu Sun, Qianfang Zou, Xin Tong, and Yang Liu.
\newblock Learning adaptive hierarchical cuboid abstractions of {3D} shape
  collections.
\newblock In {\em SIGGRAPH Asia}, 2019.

\bibitem{Tatarchenko:2018}
M. {Tatarchenko}, J. {Park}, V. {Koltun}, and Q. {Zhou}.
\newblock Tangent convolutions for dense prediction in 3d.
\newblock In {\em CVPR}, 2018.

\bibitem{Tulsiani:2017}
Shubham Tulsiani, Hao Su, Leonidas~J. Guibas, Alexei~A. Efros, and Jitendra
  Malik.
\newblock Learning shape abstractions by assembling volumetric primitives.
\newblock In {\em CVPR}, 2017.

\bibitem{Xu:2020}
Q. {Xu}, X. {Sun}, C.~Y. {Wu}, P. {Wang}, and U. {Neumann}.
\newblock {Grid-GCN} for fast and scalable point cloud learning.
\newblock In {\em CVPR}, 2020.

\bibitem{Zou:2017}
Chuhang Zou, Ersin Yumer, Jimei Yang, Duygu Ceylan, and Derek Hoiem.
\newblock {3D-PRNN}: Generating shape primitives with recurrent neural
  networks.
\newblock In {\em ICCV}, 2017.

\end{thebibliography}
}
\clearpage
 
\renewcommand{\thesection}{S}
\renewcommand{\thetable}{S\arabic{table}}
\renewcommand{\thefigure}{S\arabic{figure}}
\setcounter{table}{0}  
\setcounter{figure}{0} 

\section*{Supplementary}


In this supplementary, we first evaluate the ability of our approach to process real scan data with real noise pattern (Section~\ref{sec:evaluation_cpfn_real_scans}). Then, we visualize zoomed-in renderings of CPFN outputs to demonstrate its ability to capture smaller primitives compared to the baselines (Section~\ref{sec:evaluation_cpfn_zoom_traceparts}).  We further report the performance of CPFN in terms of memory and computational complexity (Section~\ref{sec:evaluation_cpfn_complexity}). Then, we provide an individual evaluation of both the patch selection network accuracy (Section~\ref{sec:evaluation_patch_selection}) and the merging strategy against a widely used commercial solver Gurobi \cite{gurobi} (Section~\ref{sec:evaluation_primitive_merging}).

We then further evaluate CPFN by varying the input resolution (Section~\ref{sec:evaluation_cpfn_primitive_type}) and the value of the primitive scale threshold $\eta$ (Section~\ref{sec:evaluation_cpfn_eta}). We also evaluate the impact of the value of the target number of primitives $K_{\text{glob}}$ and $K_{\text{loc}}$ on shapes with fewer primitives (Section~\ref{sec:evaluation_cpfn_primitive_scale}). Then, we report the performance of the global SPFN using a new sampling strategy based on curvature to create the low resolution point clouds (Section~\ref{sec:evaluation_spfn_importance_sampling_curvature}). We also assess alternative patch selections by sampling patches either on parts with poor detection with the global SPFN (Section~\ref{sec:evaluation_cpfn_patch_selection_global_iou}) or with high curvature areas (Section~\ref{sec:evaluation_cpfn_patch_selection_global_curvature}).

We finally provide additional details on our architecture (Section~\ref{sec:architecture_details}). Our implementation in Pytorch is publicly available on our GitHub page: \href{https://github.com/erictuanle/CPFN}{https://github.com/erictuanle/CPFN}

\subsection{Qualitative Evaluation on Real Scan Data}
\label{sec:evaluation_cpfn_real_scans}

\begin{figure}[t!]
\centering
\vspace{-0.5\baselineskip}
\newcolumntype{Y}{>{\centering\arraybackslash}X}
\scriptsize
{
\setlength{\tabcolsep}{0.0em}
\renewcommand{\arraystretch}{0.7}
\begin{tabularx}{\linewidth}{CCCCCC}
  \includegraphics[width=\linewidth]{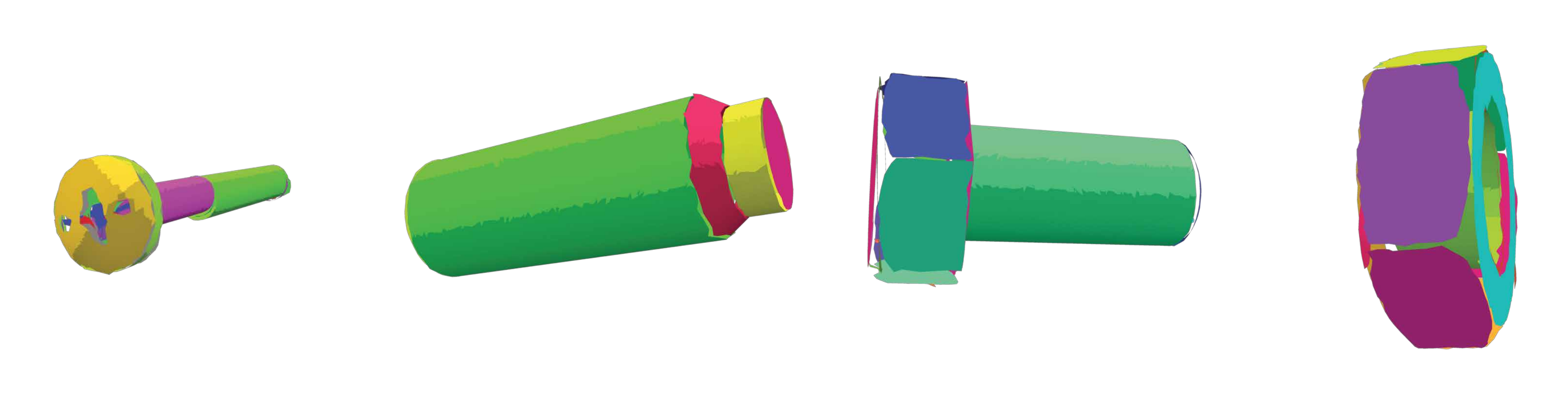}\\
\end{tabularx}
}
\vspace{-0.5\baselineskip}
\caption{CPFN results on real scans (from~\cite{SPFN}).}
\label{fig:real_scans_visu}
\vspace{-1\baselineskip}
\end{figure}

We tested our CPFN with the real scans provided by \cite{SPFN}.
The resolution of the provided scans is much lower than our synthetic dataset: only $20$k points compared to $128$k points used in our experiments. The pattern and the scale of the noise are also significantly different with our previous experiments. Yet, CPFN provides reasonable results as shown in Figure~\ref{fig:real_scans_visu}.
\subsection{Qualitative Evaluation of Small Primitive Detection and Fitting}
\label{sec:evaluation_cpfn_zoom_traceparts}

\begin{figure*}[t!]
\centering
\newcolumntype{Y}{>{\centering\arraybackslash}X}
\scriptsize
{
\setlength{\tabcolsep}{0.0em}
\renewcommand{\arraystretch}{0.9}
\begin{tabularx}{\textwidth}{m{0.08\textwidth}m{0.92\textwidth}}
  \makecell{GT\\Primitives} & \includegraphics[width=0.92\textwidth]{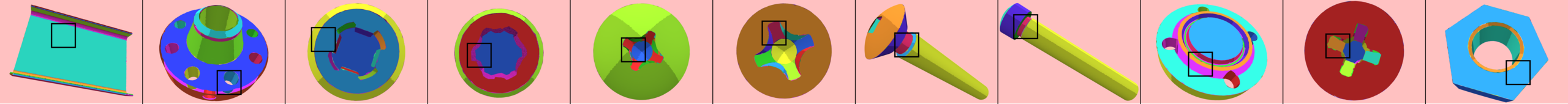} \\
  \makecell{RANSAC\\~\cite{Schnabel:2007}} & \includegraphics[width=0.92\textwidth]{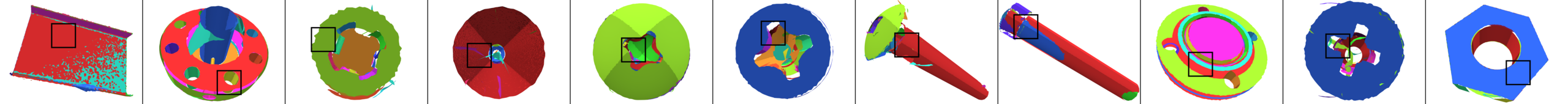} \\
  \makecell{SPFN~\cite{SPFN}} & \includegraphics[width=0.92\textwidth]{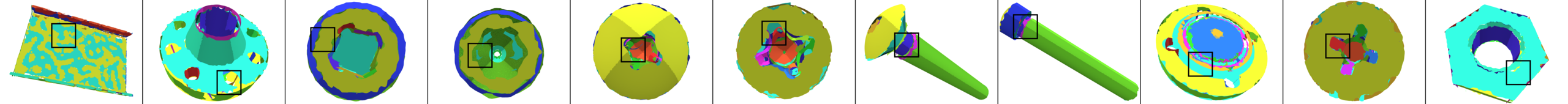} \\
  \makecell{CPFN - 5\%} & \includegraphics[width=0.92\textwidth]{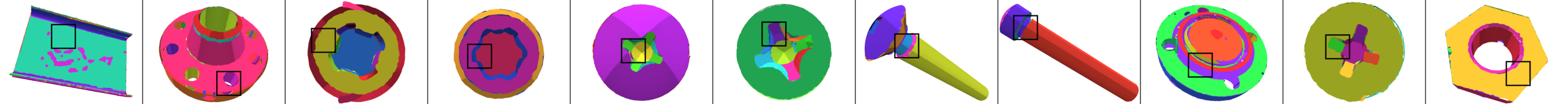} \\
  \midrule
  \makecell{GT\\Primitives} & \includegraphics[width=0.92\textwidth]{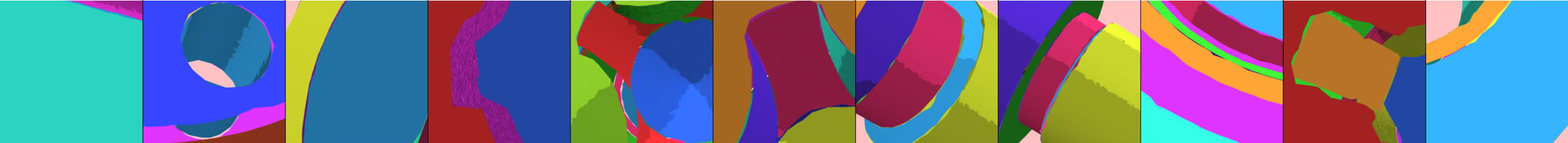} \\
  \makecell{RANSAC\\~\cite{Schnabel:2007}} & \includegraphics[width=0.92\textwidth]{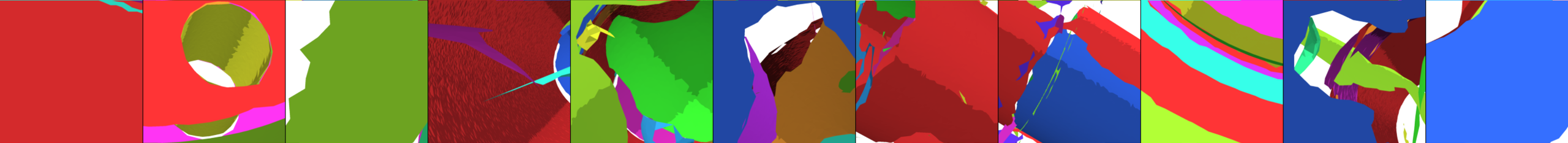} \\
  \makecell{SPFN~\cite{SPFN}} & \includegraphics[width=0.92\textwidth]{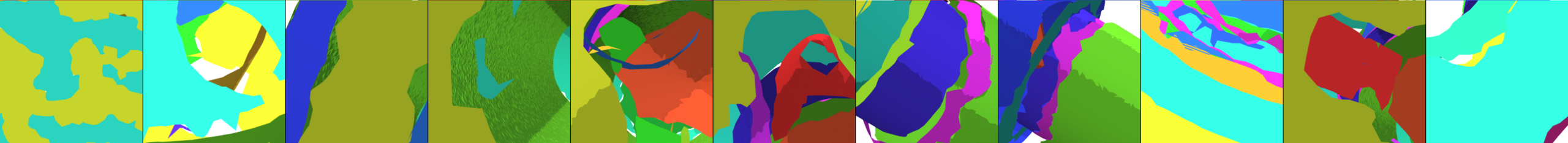} \\
  \makecell{CPFN - 5\%} & \includegraphics[width=0.92\textwidth]{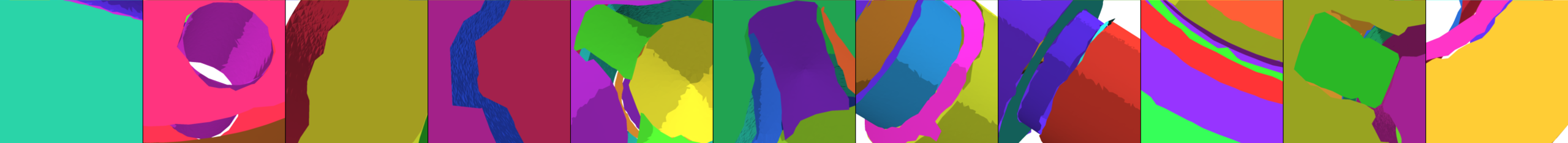} \\
\end{tabularx}
}
\vspace{-0.5\baselineskip}
\caption{Primitive fitting results for RANSAC, plain SPFN and our CPFN networks (Rows 1-4) and zoomed-in visualizations on smaller primitives (Rows 5-8). RANSAC and SPFN directly operate on the global object failing on the small primitives. Our CPFN pipeline estimates the heatmaps corresponding to small primitives at different scales and learns a better primitive decomposition on local patches sampled from such regions improving the detection of small primitives.}
\label{fig:zoomin_results}
\vspace{-1\baselineskip}
\end{figure*}

To visually assess the efficiency of CPFN in detecting small primitives, we visualize in Figure~\ref{fig:zoomin_results} zoomed-in renderings of the predictions for smaller primitives. As shown, our CPFN pipeline improves significantly the detection and the fitting of the smaller primitives compared to the RANSAC~\cite{Schnabel:2007} and SPFN~\cite{SPFN} baselines.
\subsection{Memory and Computation Complexity}
\label{sec:evaluation_cpfn_complexity}

To run SPFN~\cite{SPFN} at training time, one would need to double the average consumer GPU memory for a batch size of 8 for the high resolution Traceparts~\cite{TraceParts} dataset while our method could still be applicable for even higher resolutions.

At test time, CPFN only requires three forward passes - through the patch selection network (${\sim}0.02$s) and the global (${\sim}0.25$s) and local (${\sim}0.22$s) SPFN respectively - followed by the merging step (${\sim}0.43$s). Note that our heuristic merging step takes less than one second per scan, which is incomparably faster than running a commercial optimization solver as described in Section~\ref{sec:evaluation_primitive_merging}. Both SPFN~\cite{SPFN} and CPFN have similar memory footprint at test time.

%
%
\subsection{Quantitative Evaluation of the Patch Selection Network}
\label{sec:evaluation_patch_selection}

%
%

We assess the accuracy, precision and recall of our patch selection network trained to identify the regions which are likely to contain small primitives at test time. We report the quantitative results in Table~\ref{tab:benchmark_patch_selector} for varying scales of primitives, i.e., $\eta\in\{1\%,2\%,3\%,4\%,5\%\}$ (see Section~\ref{sec:patchSelection}).

By increasing the value of $\eta$, we enlarge the set of primitives considered as small and as such, we also increase the number of local patches to retrieve. The accuracy of our patch selection network varies between $87.96\%$ and $98.46\%$ but slightly decreases with the scale. The average precision and recall is $57.21\%$ and $52.75\%$ respectively emphasizing that identifying fine-scale regions in a low-resolution point cloud is a challenging task. Nevertheless, our pipeline is not that sensitive to the accuracy of the patch selection stage. In fact, in our ablation studies (Table~\ref{tab:results_traceparts}, row 14 in the paper), we show that a perfect patch selection has a marginal impact on the output performance.

%
\begin{table}[h!]
\footnotesize{
\setlength{\tabcolsep}{2.0pt}
  \centering
  \caption{Accuracy and precision/recall trade-off of our patch selection network trained to identify small primitives, i.e. primitives with less than $\eta\cdot N$ points (see Section~\ref{sec:patchSelection}).}
  \vspace{0.3\baselineskip}
   
    \begin{tabularx}{\linewidth}{c|C|C|C|C|C|C|C}
    \toprule
    Scale $\eta$ & Accu. & Prec. & Recall & TP & FP & TN & FN \\
    \midrule
    1\%  & \textbf{98.46} & \textbf{60.85} & 37.14 &           0.66  &           \textbf{0.43}  &        \textbf{97.79}  &            \textbf{1.12}  \\
    2\%  & 96.27 & 57.65 & 52.82 &           2.29  &           1.68  &        93.98  &            2.05  \\
    3\%  & 91.98 & 54.65 & 53.25 &        4.70  &        3.90  &        87.28  &         4.12  \\
    4\%  & 88.99 & 54.23 & 59.35 &        7.20  &        6.08  &        81.79  &         4.93  \\
    5\%  & 87.96 & 58.65 & \textbf{61.21} &        \textbf{8.99}  &        6.34  &        78.79  &         5.70  \\
    \bottomrule
    \end{tabularx}
  \label{tab:benchmark_patch_selector}
}
\vspace{-\baselineskip}
\end{table}
%
\subsection{Quantitative Evaluation of our Merging Heuristic Compared to a Binary Programming Solver}
\label{sec:evaluation_primitive_merging}

As explained in Section~\ref{sec:patchMerging}, our segment merging problem is a quadratic binary programming problem. Thus, instead of using our Hungarian-algorithm-based merging method, we tried a commercial solver, Gurobi~\cite{gurobi} (using a branch-and-bound algorithm), and compared the performance with ours in terms of both the computation time and the segmentation accuracy after the merging. Since the branch-and-bound method may take a huge amount of time, we set the time limit to 10 mins. When we tested both methods with 26 randomly picked test cases, in 22 cases out of them, Gurobi failed to find the optimum solution in 10 mins. Contrarily, our method took $0.43$ secs on average. The resulting mIoU of Gurobi for the random 26 cases was $50.54\%$, while our heuristic achieved $81.42\%$. This gap in performance highlights that even after 10 mins, the last solution found by Gurobi is still far away from optimum. Even when Gurobi managed to get a better solution than ours (with a lower energy in the optimization) before the 10 mins limit, the improvement on mIoU was marginal: $+0.09\%$.

\subsection{Quantitative Evaluation of CPFN with Different Point Cloud Resolutions}
\label{sec:evaluation_cpfn_primitive_type}

We report the per-primitive-type mIoU in Table~\ref{tab:results_type_miou} at two different input resolutions - 16k and 128k. The comparison between the results of SPFN \cite{SPFN} and our CPFN  at 16k and 128k shows that higher resolutions take more benefits from our two-level prediction architecture.

Also CPFN performs better than SPFN~\cite{SPFN} for all primitive types. The gap between the highest (Sphere) and the lowest (Cone) mIoUs is also smaller compared to SPFN~\cite{SPFN}, meaning that our performance is more balanced across the different primitive types.
%
\begin{table}[h!]
\footnotesize{
\setlength{\tabcolsep}{2.0pt}
  \centering
  \caption{Percentages of each primitive type (top row) and mean IoUs of SPFN \cite{SPFN} and CPFN for each type at 16k and 128K input resolutions.}
  \vspace{0.3\baselineskip}
    \begin{tabularx}{\linewidth}{c|c|CCCC}
    \toprule
    \multicolumn{2}{c|}{Type} & Cone & Cylinder & Plane & Sphere \\
    \midrule
    \multicolumn{2}{c|}{Pct. Primitives} & 26.95 & 29.23 & 39.98 & 3.84 \\
    \midrule
    \multirow{2}{*}{16k} & SPFN & 56.51 & 67.29 & 66.11 & 82.70\\
    & CPFN & 68.85 & 75.94 & 74.83 & 86.68 \\
    \midrule
    \multirow{2}{*}{128k} & SPFN & 57.13 & 67.54 & 66.38 & 83.24\\
    & CPFN & \textbf{76.46} & \textbf{77.85} & \textbf{80.57} & \textbf{87.79} \\
    \bottomrule
    \end{tabularx}
  \label{tab:results_type_miou}
}
\vspace{-\baselineskip}
\end{table}
%
\subsection{Impact of the Primitive Scale Threshold $\eta$}
\label{sec:evaluation_cpfn_eta}

We analyze the effect of an increase for the value of the primitive scale threshold $\eta$ (see Section~\ref{sec:patchSelection}), which is the maximum scale that can be selected by our patch selection network, in Table~\ref{tab:variation_eta}. Doubling $\eta$ from $5\%$ to $10\%$ brings a slight increase in the performance for the bigger primitives but reduces the ability to detect small primitives.

%
\begin{table}[h!]
\footnotesize{
\setlength{\tabcolsep}{2.0pt}
  \centering
  \caption{Segmentation accuracy of CPFN at various primitive scales with two different $\eta$ values $5\%$ and $10\%$. Each scale bucket contains roughly the same number of primitives}
  \vspace{0.3\baselineskip}
    \begin{tabularx}{\linewidth}{m{1.8cm}|CCCCC}
    \toprule
    Scale & $\sim$1\% & 1\%$\sim$2\% & 2\%$\sim$4\% & 4\%$\sim$12\% & 12\%$\sim$ \\
    \midrule
    RANSAC~\cite{CgalRANSAC} & 34.68 & 40.38 & 56.78 & 70.63 & 69.50 \\
    SPFN~\cite{SPFN}  & 44.25 & 55.53 & 70.12 & 74.29 & 79.75 \\
    CPFN - 5\% & \textbf{65.74} & 77.31 & 84.19 & 83.55 & 83.95 \\
    CPFN - 10\% & 65.58 & \textbf{77.49} & \textbf{84.31} & \textbf{85.16} & \textbf{84.66}\\
    \bottomrule
    \end{tabularx}
  \label{tab:variation_eta}
}
\vspace{-\baselineskip}
\end{table}
%
\subsection{Impact of the Values of the Maximum Number of Primitives Predicted by each SPFN $K_{\text{glob}}$ and $K_{\text{loc}}$}
\label{sec:evaluation_cpfn_primitive_scale}

We set reasonably big numbers for $K_{\text{glob}}$ and $K_{\text{loc}}$ (see Section~\ref{sec:patchMerging}) to handle all test cases in our dataset. But, obviously we cannot set very huge numbers due to the GPU memory issue --- many neural networks including SPFN~\cite{SPFN} have the same technical issue. If the input scan has a huge number of primitives, as an alternative, we can consider using only local SPFN with small size patches. Table~\ref{tab:impact_hyperameters} shows how the performance of our CPFN changes when the number of primitives in an input varies for the fixed values of $K_{\text{glob}}=28$ and $K_{\text{loc}}=21$. One worthwhile observation is that setting big numbers to $K_{\text{glob}}$ and $K_{\text{loc}}$ does not affect the case of having a few primitives in the input scan
%
\begin{table}[h!]
\footnotesize{
\setlength{\tabcolsep}{2.0pt}
  \centering
  \caption{
  Segmentation accuracy (\%) of CPFN with respect to the number of primitives in the input shape.
  }
  \vspace{0.3\baselineskip}
    \begin{tabularx}{\linewidth}{>{\centering}m{2.0cm}|CCCCC}
    \toprule
    Nb Primitives & 3$\sim$8 & 8$\sim$12 & 12$\sim$14 & 14$\sim$25 \\
    \midrule
    CPFN ($\eta=5\%$) & 78.58 & 81.36 & 83.17 & 74.61 \\
    \bottomrule
    \end{tabularx}
  \label{tab:impact_hyperameters}
}
\vspace{-\baselineskip}
\end{table}
%
\subsection{Curvature-based Importance Sampling for the Global SPFN}
\label{sec:evaluation_spfn_importance_sampling_curvature}

Instead of generating the low-resolution point cloud with FPS sampling only, we evaluate a new importance sampling strategy based on curvature. We first estimated point curvatures directly from the point cloud via jet-fitting using CGAL~\cite{CgalRANSAC} with default parameters.
(We clipped outlier values due to numerical instabilities.)
We trained SPFN using low-resolution point clouds sampled in two steps: (i) half of the points were sampled using FPS sampling, and (ii) the remaining half was randomly sampled proportionally to mean curvature. For the Traceparts~\cite{TraceParts} dataset, we obtain an mIoU of $\mathbf{65.49\%}$, which is lower than $\mathbf{66.29\%}$ with the full FPS based sampling and also $\mathbf{79.64\%}$ with our CPFN pipeline.
\subsection{Patch Selection Based on Global SPFN IoUs}
\label{sec:evaluation_cpfn_patch_selection_global_iou}

\begin{figure*}[t]
\centering
\newcolumntype{Y}{>{\centering\arraybackslash}X}
\scriptsize
{
\setlength{\tabcolsep}{0.0em}
\renewcommand{\arraystretch}{0.9}
\begin{tabularx}{\textwidth}{m{0.08\textwidth}m{0.92\textwidth}}
  \makecell{1-mIoU} & \includegraphics[width=0.92\textwidth]{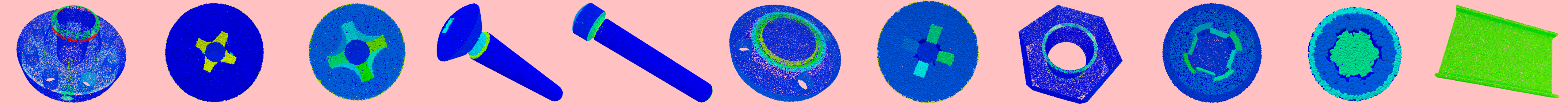} \\
  \makecell{Scale $< \eta$} &
  \includegraphics[width=0.92\textwidth]{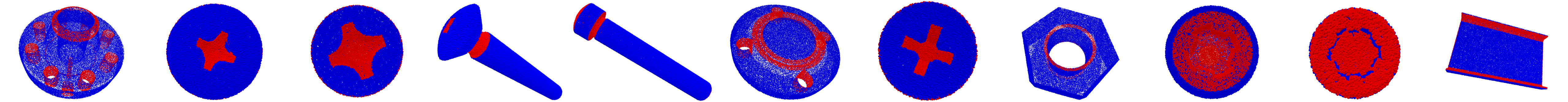} \\
\end{tabularx}
}
\vspace{-0.5\baselineskip}
\caption{Comparison between the heatmap of global SPFN IoUs and the heatmap highlighting small primitive areas with $\eta=5\%$. Both heatmaps highlight similar areas for most of the objects, showing that sampling patches from either heatmap will produce similar patch samples. Heatmaps are displayed with the \textit{Jet} color map going from blue to green to red.}
\label{fig:heatmap}
\vspace{-1\baselineskip}
\end{figure*}

In order to assess our sampling strategy described in Section~\ref{sec:patchSelection}, we compare two types of heatmaps: (i) the heatmap highlighting areas with low global SPFN mIoU and (ii) the heatmap estimated by our patch selection network to identify small primitives.
As shown in Figure~\ref{fig:heatmap}, both heatmaps have high values in similar areas, meaning that sampling patches from either heatmap will produce similar patch samples.
The last column in the figure (column 11) is an exceptional case when the global SPFN fails to properly segment large primitives due to their proximity.
\subsection{Patch Selection Based on Curvature}
\label{sec:evaluation_cpfn_patch_selection_global_curvature}

We also tried estimating the point heatmaps based on the previously computed mean curvature. We sampled patches from the top 20\% points with the highest mean curvature at training and test time.
This approach did not perform as good as our original CPFN pipeline:
$\mathbf{76.00\%}$ mIoU compared with $\mathbf{79.64\%}$ of ours as high curvature areas do not cover all of the small primitives.

\subsection{Implementation and Training Details}
\label{sec:architecture_details}

We provide a more detailed description of our architecture and parameters used in our experiments. We first explain the design of both the global and local SPFN in Section~\ref{sec:architecture_spfn} and the patch selection network in Section~\ref{sec:architecture_patch_selection_network}. We then detail how we pre-processed the dataset (Section~\ref{sec:architecture_dataset}) and the optimization parameters used for the training (Section~\ref{sec:architecture_optimization}). We finally explain how the full pipeline operates at training (Section~\ref{sec:architecture_training}) and test time (Section~\ref{sec:architecture_test}).
%
%

\subsubsection{Supervised Primitive Fitting Network (SPFN)}
\label{sec:architecture_spfn}

Throughout our experiments, we use our re-implementation of the SPFN~\cite{SPFN} pipeline in Pytorch.

SPFN~\cite{SPFN} is itself based on a single-scale PointNet++~\cite{PointNet++} for which the default hyperparameters are used. The default PointNet++ implementation is designed as an encoder-decoder architecture, which progressively decreases the point cloud resolution with depth, from the input resolution to $512$, $128$ and finally to a single point vector. The decoder symmetrically upsamples the point cloud to the input resolution.
After a common linear layer,
the last PointNet++~\cite{PointNet++} layers is replaced to produce three per-point outputs for the segmentation $\mathbf{W}$, the type $\mathbf{T}$ and the normal $\mathbf{N}$ from three dense layer heads. On top of PointNet++ backbone, SPFN~\cite{SPFN} (see Section~\ref{sec:globalSPFN}) computes the primitive parameters based on the network outputs and supervises the training on five different losses: (i)~segmentation loss $\mathcal{L}_{\text{seg}}$, (ii)~normal loss $\mathcal{L}_{\text{norm}}$, (iii)~primitive type loss $\mathcal{L}_{\text{type}}$, (iv)~residual loss $\mathcal{L}_{\text{res}}$, i.e., fitting loss, and (v)~axis loss $\mathcal{L}_{\text{axis}}$.

We keep unchanged almost all implementation details from the original Tensorflow implementation. All hyperparameters are kept identical for both of our SPFN-based sub-networks, i.e., global (Section~\ref{sec:globalSPFN}) and local SPFN (Section~\ref{sec:patchPrediction}). We train separately the global and local SPFN since they act on different scales of the input point cloud. As explained in the paper, we  modify the local SPFN to use the global $\mathbf{l}_o$ and local context $\mathbf{l}_i^g$ extracted from the global SPFN by feeding an augmented latent code $\mathbf{l}_i^{'} = \left[\mathbf{l}_i, \quad \mathbf{l}_o, \quad \mathbf{l}_i^g\right]$ to the decoder.
The first module from the decoder is thus modified to accept the additional input channels.
Another difference is the maximum number of primitives each SPFN instance can predict. As global objects are likely to contain more primitives than patches, the global SPFN is trained to be able to predict more primitives ($K_{\text{glob}}$) than the local SPFN counterpart ($K_{\text{loc}}$). 
This boils down to change the number of output channels of the segmentation head to the desired number of primitives
. For the TraceParts~\cite{TraceParts} dataset, we fix $K_{\text{glob}}=28$ and $K_{\text{loc}}=21$ which are respectively the maximum number of primitives in a single object and in a single patch found in the dataset.

Both SPFNs are trained using ground truth information extracted from the CAD models, either at the object- or the patch-level: (i) point-to-primitive assignment, (ii) point normals, (iii) primitive types and (iv) primitive axis (except for spheres).

\subsubsection{Patch Selection Network}
\label{sec:architecture_patch_selection_network}

The patch selection network (Section~\ref{sec:patchSelection}), also based on the same default PointNet++~\cite{PointNet++} implementation, produces a single binary classification tensor and is trained using the binary cross-entropy loss.
Thus, we replace the last linear layer of the backbone PointNet++~\cite{PointNet++} by a new one with 2 output channels.
The GT small-primitive heatmaps for supervision are based on the primitives which area are smaller than a threshold $\eta$.

\subsubsection{Dataset Parameters}
\label{sec:architecture_dataset}

The Traceparts~\cite{TraceParts} dataset is pre-processed to merge adjacent primitives  sharing  the  same  parameters and  discard  extremely tiny primitives. Differently to \cite{SPFN}, only primitives with an area smaller than $0.5\%$ (instead of $2\%$) of the entire object are discarded to make sure smaller primitives are included in the dataset. Higher resolution point clouds allow our approach to capture accurately small primitives that were originally discarded. The point clouds are
normalized to the unit sphere and
randomly perturbed with uniform noise [$-5e-3$, $5e-3$] along the ground truth normal direction. For comparison, the average sampling distance ($d_s\approx 5e-3$) is equivalent to that noise level.

\subsubsection{Optimization Parameters}
\label{sec:architecture_optimization}

For all SPFN modules, we use a batch size of $16$ samples with both (i) a learning decay (initial learning rate $10^{-3}$ with staircase learning decay $0.7$) and (ii) a batch norm decay (initial batch norm decay $0.5$ with staircase decay $0.5$). For optimization, we use Adam~\cite{adam} with $\beta$ coefficients of $0.9$ and $0.999$ for the gradient and its square respectively. For the patch selection network, the same set of training parameters are used except the batch size which is increased to $32$.

The global SPFN and the patch selection network are trained with 100 epochs. The local SPFN input dataset is much bigger than the original dataset as multiple patches will be sampled on the same object. Thus, we fix the number of epochs for the local SPFN so that all trainings have the same number of iterations.

\subsubsection{CPFN Pipeline at Training Time}
\label{sec:architecture_training}

Our full CPFN pipeline is trained as a sequential cascaded process. We first train the global SPFN (Section~\ref{sec:globalSPFN}) and the patch selection network (Section~\ref{sec:patchSelection}) on downsampled point clouds $n=8\,192$. The output of the global SPFN provides an initial primitive decomposition of the input object with poor accuracy on small primitives. The patch selection network is thus trained to learn how to sample patches in those smaller primitive scale areas. The patch selection network is only used at test time when primitive information is not available. To improve on the initial global SPFN outputs, we train in turn our local SPFN (Section~\ref{sec:patchPrediction}) on patches of $n=8\,192$ points randomly samples in GT small primitive areas. The patches are sampled sequentially to cover as much as possible all small primitives. At each iteration, a new patch is sampled on small primitives from a random query point that has not yet been covered by any of the previous patches. We limit the maximum number of patches for a given object to $32$. To provide both local and global context to this local SPFN, we augment the patch latent vector with the object latent vector $\mathbf{l}_o$ and the patch centroid's features $\mathbf{l}_i^g$, both extracted via the trained global SPFN.

\subsubsection{CPFN Pipeline at Test Time}
\label{sec:architecture_test}

At test time, we run in parallel the global SPFN and the patch selection network to generate respectively the initial primitive decomposition and the small primitive heatmap. Contrary to the training, the global SPFN is tested on the high-resolution version of the point cloud. However, the patch selection network is still tested on the low-resolution point cloud. The heatmap - produced by the patch selection network - generates values in $[0,1]$ for each of the $n=8\,192$ points - higher values meaning higher chances for the point to be part of a small primitive. Then patches are randomly sampled in areas where the predicted value is above $\theta=0.5$ so that all such points are covered by at least one patch. The local SPFN is then run on those newly sampled patches to refine the fitting on the smaller primitives. Finally, the segments from both the global and the local SPFNs are merged to produce the final primitive decomposition with improved performance on small primitives by following the process explained in Section~\ref{sec:patchMerging} of the paper.

\end{document}